\documentclass{article}

\PassOptionsToPackage{numbers, compress}{natbib}
\usepackage{arxiv}

\usepackage[utf8]{inputenc} % allow utf-8 input
\usepackage[T1]{fontenc}    % use 8-bit T1 fonts
\usepackage{hyperref}       % hyperlinks
\usepackage{url}            % simple URL typesetting
\usepackage{booktabs}       % professional-quality tables
\usepackage{amsfonts}       % blackboard math symbols
\usepackage{nicefrac}       % compact symbols for 1/2, etc.
\usepackage{microtype}      % microtypography
\usepackage{xcolor}         % colors

\usepackage{xspace}
\usepackage{amsmath}
\usepackage{graphicx}
\usepackage{wrapfig}
\usepackage{multirow}
\usepackage{makecell}
\usepackage{booktabs}

\usepackage{tabularx}
\usepackage{array}
\usepackage{ragged2e}
\usepackage{enumitem} 
\usepackage{listings}
\usepackage{amssymb} % for \checkmark
\usepackage[most]{tcolorbox}
\usepackage{natbib}

\newcommand{\dataset}[1]{\text{\textsc{#1}}\xspace} 

\newcommand{\model}[1]{\textsc{#1}\xspace}

\newcommand{\pipeline}[1]{\text{\texttt{#1}}\xspace} 

\title{EO-Gym: A Multimodal, Interactive Environment for Earth Observation Agents}

\author{
  \textbf{Sai Ma}$^{1}$ \quad
  \textbf{Zhuang Li}$^{2}$ \quad
  \textbf{Sichao Li}$^{3}$ \quad
  \textbf{Xinyue Xu}$^{4}$ \\
  \textbf{Ruibiao Zhu}$^{1}$ \quad
  \textbf{Tony Boston}$^{1}$ \quad
  \textbf{John A. Taylor}$^{1}$ \\
  \\
  $^{1}$The Australian National University \\
  $^{2}$Royal Melbourne Institute of Technology \\
  $^{3}$University of Sydney \\
  $^{4}$The Hong Kong University of Science and Technology \\
  \\
  \texttt{\{sai.ma, ruibiao.zhu, tony.boston, john.taylor\}@anu.edu.au} \\
  \texttt{zhuang.li@rmit.edu.au}, \texttt{sichao.li@sydney.edu.au}, \texttt{xinyue.xu@connect.ust.hk}
}

\begin{document}

\maketitle

\begin{abstract}

Earth Observation (EO) analysis is inherently interactive: resolving uncertainty often requires expanding the region of interest, retrieving historical observations, and switching across sensors such as optical and Synthetic Aperture Radar. However, most EO benchmarks collapse this process into fixed-input, single-turn tasks. To address this gap, we present \pipeline{EO-Gym}, a controlled executable framework for multimodal, tool-using EO agents that formulates EO analysis as a Gymnasium-style local geospatial workspace backed by more than $660$k multimodal files indexed by location, time, and sensor type, with $35$ EO-specialized tools spanning six task families. Built on this environment, we construct \dataset{EO-Gym-Data}, a benchmark of $9{,}078$ trajectories and $34{,}604$ reasoning steps, and grounded in eight public EO datasets together with Landsat and Sentinel-2 imagery. Evaluating $10$ open and closed VLMs shows that strong general-purpose models still struggle with interactive EO reasoning, especially on temporal and cross-modal workflows. As a reference baseline, \model{EO-Gym-4B}, obtained by fine-tuning \model{Qwen3-VL-4B-Instruct} on \dataset{EO-Gym-Data}, improves overall Pass@3 from $0.49$ to $0.74$ under the main evaluation setting. \pipeline{EO-Gym} provides a reproducible environment for interactive EO agents, operationalizing EO as an evidence-gathering problem that requires planning across geospatial, temporal, and sensing modality. Code and data are available at \url{https://huggingface.co/datasets/paperuploadacount/EO-Gym}.

\end{abstract}

\section{Introduction}
Earth Observation (EO) underpins planetary-scale monitoring, disaster response, and environmental change analysis. However, many EO questions cannot be resolved from fixed observations. Analysts can reduce uncertainty by expanding the region of interest, retrieving historical observations, changing spatial scale, or switching sensing modalities, for example, from optical imagery to Synthetic Aperture Radar (SAR) during periods of cloud cover or at night \citep{joshi2016opticalsar,claverie2018hls,ju2025hlsv2}. Moreover, because decades of prior work remain siloed within current operational frameworks, accessing this historical data requires systems capable of automatically interfacing with disparate legacy architectures \citep{wulder2016landsat}. Consequently, EO analysis is not merely a perception problem, but an interactive evidence-acquisition process that requires stateful tools to navigate space, reason over time, switch modalities, and synthesize heterogeneous observations for geospatial decision-making.

Recent progress in Vision Language Models (VLMs) and tool-using agents presents a promising avenue for automating these analytical workflows \citep{liu2024llavaplus,yao2023react,qin2024toolllm,koh2024visualwebarena,xie2024osworld}. However, the dominant large-scale evaluation paradigms in EO remain fundamentally static, as existing corpora are restricted to fixed-input comprehension rather than dynamic, sequential evidence gathering \citep{zhang2024earthgpt,zhan2025skyeyegpt,li2024vrsbench,lacoste2023geobench,soni2025earthdial}. While emerging EO agent initiatives have started to address this limitation \citep{kao2025univearth,shabbir2025thinkgeo,feng2026earthagent,shabbir2026openearthagent,zhao2026openearthagent}, the community critically lacks a controlled and reproducible multimodal benchmark. To effectively support the interactive, cross-modal evidence acquisition required for planetary-scale analysis, advancing this domain requires a comprehensive framework. This environment must simultaneously provide executable interactions across space and time, as well as scalable trajectory generation for training and evaluation.

To address these requirements, we introduce \pipeline{EO-Gym}, a Gymnasium-style workspace designed to directly support executable interactions across EO tasks \citep{brockman2016gym}. Backed by over $660$k multimodal indexed files and $35$ specialized tools, this environment allows agents to perform complex EO reasoning by dynamically accessing data to reduce uncertainty. Utilizing this framework, we ensure scalable trajectory generation by constructing \dataset{EO-Gym-Data}. This comprehensive benchmark comprises $9,078$ trajectories derived from diverse satellite sensors, spanning true-color, multispectral, and SAR imagery \citep{wulder2016landsat,drusch2012sentinel,torres2012sentinel1,digitalglobe2014worldview3,toutin2002quickbird,madden2009geoeye1,huang2018gf2,wmo2026jilin1,cyclomedia2026aerial}. Moving beyond static predictions, each trajectory moves beyond static input by linking tool calls, observations, and reasoning into an explicit path for interactive temporal, geospatial, and cross-modal evidence gathering.

To assess whether current models can gather temporal, geospatial, and cross-modal evidence directly from an EO environment, we evaluate 10 open and closed VLMs in \pipeline{EO-Gym} and find that general-purpose models struggle with these dynamic workflows. We then fine-tune \model{Qwen3-VL-4B-Instruct} to obtain \model{EO-Gym-4B}, which improves Pass@3 from $0.49$ to $0.74$ under the Verified/Skill/Simple protocol. This gain shows that training in an interactive EO environment substantially improves tool use and core EO reasoning.

In summary, our contributions are threefold:
\begin{itemize}
    \item \textbf{Controlled executable EO environment.} We introduce \pipeline{EO-Gym}, a multimodal geospatial workspace that supports spatial navigation, temporal retrieval, cross-modal switching, and EO-specific tool use in a fully executable local setting.
    \item \textbf{EO multimodal trajectory benchmark.} We construct \dataset{EO-Gym-Data}, a benchmark of $9{,}078$ EO tool-use trajectories across six EO tasks, grounded in eight public EO datasets and the EO-Gym data-gathering execution space.
    \item \textbf{Unified benchmark and reference baseline.} We evaluate $10$ open and closed VLMs and provide \model{EO-Gym-4B} as a strong EO-specialized reference baseline.
\end{itemize}

\section{Related Work}

\paragraph{General benchmarks for tool-using agents.}
Agent benchmarking increasingly relies on executable environments rather than static prompts. General tool-use benchmarks such as \dataset{ToolBench}, \dataset{GAIA}, and \dataset{GTA} evaluate tool selection, invocation, and coordination across realistic tasks \citep{qin2024toolllm,mialon2024gaia,wang2024gta}. Recent Gym-style suites extend this paradigm: \pipeline{MLGym} frames AI research as episodic agent tasks, while \pipeline{R2E-Gym} provides executable software-engineering environments with verifiable outcomes \citep{nathani2025mlgym,jain2025r2egym}. Interactive benchmarks such as \pipeline{VisualWebArena}, \pipeline{WebArena}, \pipeline{OSWorld}, and \pipeline{FormFactory} evaluate sequential behavior in web, GUI, form-filling, and computer-use settings \citep{koh2024visualwebarena,zhou2024webarena,xie2024osworld,li2025formfactory}. Related simulators such as \pipeline{AerialGym} support Gym-like aerial perception and control, but target embodied UAV learning rather than satellite catalog reasoning \citep{kulkarni2025aerialgym}. Collectively, these works show that realistic agent evaluation requires stateful, executable interaction. However, they do not address EO, where reasoning requires evidence acquisition across space, time, and modality, and therefore lacks the specialized tools and environmental grounding needed for EO workflows.

%\paragraph{General benchmarks for tool-using agents.}
%Agent benchmarking increasingly uses executable environments rather than static prompts. General tool-use benchmarks such as \dataset{ToolBench}, \dataset{GAIA}, and \dataset{GTA} evaluate tool selection, invocation, and coordination across realistic tasks \citep{qin2024toolllm,mialon2024gaia,wang2024gta}. Newer Gym-style suites extend this direction: \pipeline{MLGym} frames AI research as episodic agent tasks, and \pipeline{R2E-Gym} provides executable software-engineering environments with verifiable outcomes \citep{nathani2025mlgym,jain2025r2egym}. Interactive environments such as \pipeline{VisualWebArena}, \pipeline{WebArena}, \pipeline{OSWorld}, and \pipeline{FormFactory} further evaluate sequential behavior in web, GUI, form-filling, and computer-use settings \citep{koh2024visualwebarena,zhou2024webarena,xie2024osworld,li2025formfactory}. Related simulators such as \pipeline{AerialGym} demonstrate Gym-like interaction for aerial perception and control, but target embodied UAV learning rather than satellite catalog reasoning \citep{kulkarni2025aerialgym}. Together, these works show that realistic agent evaluation requires stateful, executable interaction. However, they do not address EO, where reasoning is an evidence-acquisition problem across space, time, and modality. General agents therefore lack the specialized tools and environmental grounding needed for EO workflows.

\paragraph{From static EO benchmarks to interactive agents.}
The EO community has developed extensive datasets for perception and vision-language learning, including \dataset{xView}, \dataset{DOTA}, \dataset{DIOR}, \dataset{FAIR1M}, \dataset{SARDet-100K}, and \dataset{VRSBench} \citep{lam2018xview,xia2018dota,li2020dior,sun2022fair1m,li2024sardet100k,li2024vrsbench}. Recent multimodal resources such as \dataset{EarthScape}, \dataset{EarthGPT}, \dataset{SkyEyeGPT}, and \dataset{EarthDial} further expand EO supervision across multi-scale geospatial analysis, multi-sensor comprehension, and dialogue-style learning \citep{massey2025earthscape,zhang2024earthgpt,zhan2025skyeyegpt,soni2025earthdial}. In parallel, infrastructure and representation-learning work has improved reusable EO data access and fixed-input reasoning: \pipeline{TorchGeo} supports geospatial datasets, samplers, transforms, and pretrained models; \pipeline{CSP} learns geospatial-visual representations from image-location contrast; and satellite time-series methods model temporal attention, panoptic sequence structure, and explainability for crop and parcel analysis \citep{stewart2022torchgeo,mai2023csp,russwurm2020selfattention,garnot2021panoptic,abbas2023xai4eo}. These resources are important for pretraining and specialist EO perception, but remain centered on fixed-input prediction rather than autonomous environmental interaction. Recent agentic EO systems address this gap along complementary axes. \citet{kao2025univearth} study EO agents through code generation and API execution. \dataset{ThinkGeo} introduces structured tool use and multi-step planning \citep{shabbir2025thinkgeo}. \dataset{Earth-Bench} cover optical and spectral tools with trajectory-level assessment \citep{feng2026earthagent}. \pipeline{OpenEarthAgent} emphasizes supervised trajectory learning and deterministic replay validation \citep{shabbir2026openearthagent}, while \pipeline{OpenEarth-Agent} targets open-environment workflow planning and tool creation \citep{zhao2026openearthagent}. However, existing systems do not provide a controlled VLM evaluation environment that jointly supports spatial expansion, temporal retrieval, and cross-modal switching, limiting their ability to assess agents that resolve uncertainty beyond the initial EO data input.

\section{Controlled Executable Earth Observation Environment: \pipeline{EO-Gym}}
\label{data-tool-and-source}

\begin{figure*}[t]
    \centering
    \includegraphics[width=\textwidth]{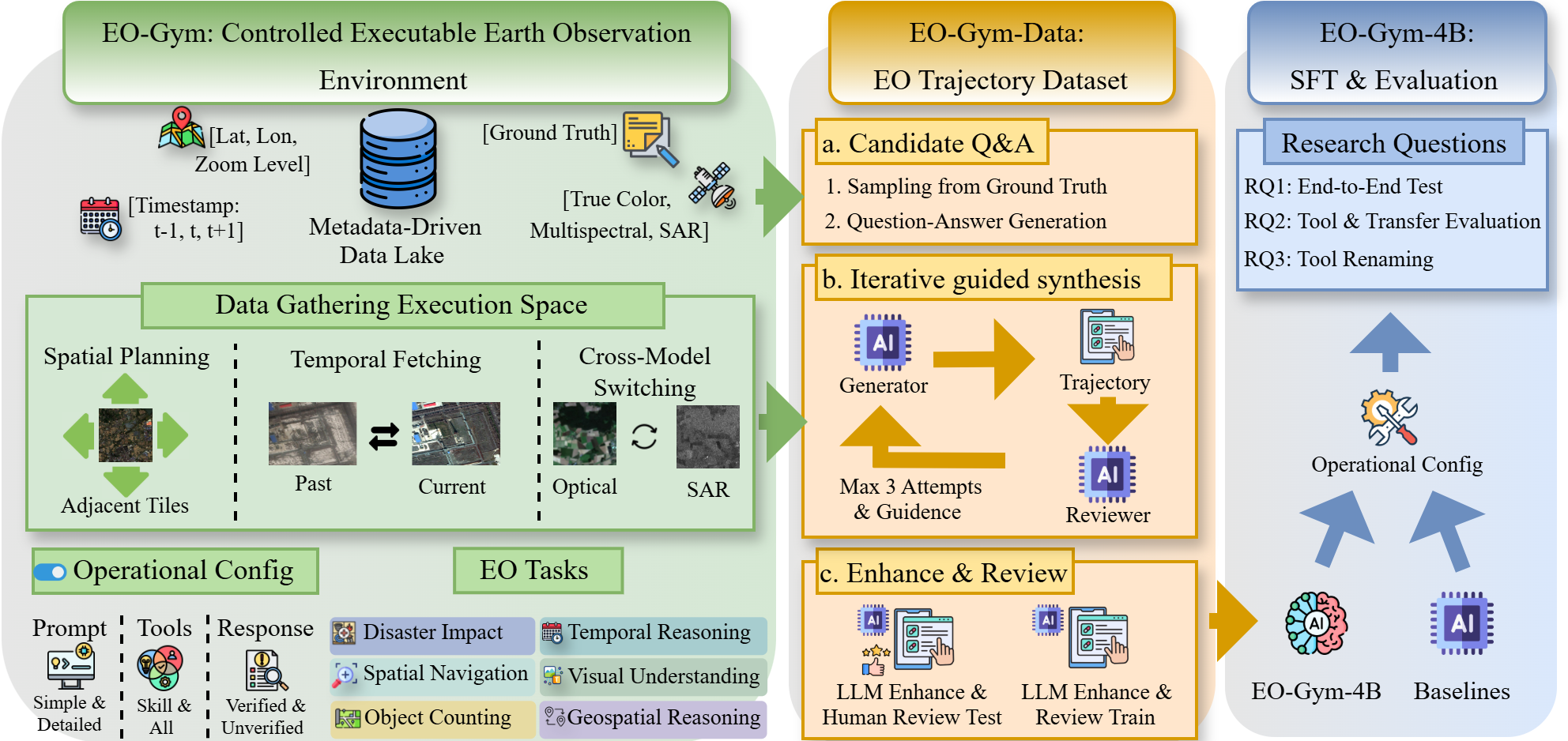}
    \caption{The \pipeline{EO-Gym} framework operationalizes EO as an interactive evidence-acquisition problem. A controlled environment (left) enables dynamic exploration across space, time, and modality. An iterative pipeline (middle) synthesizes a large-scale trajectory dataset. Finally, a unified evaluation protocol (right) assesses fine-tuned and baseline models.}
    \label{fig:eo_agent_framework}
\end{figure*}

\paragraph{Metadata-Driven Data Lake.}
The left panel of Figure~\ref{fig:eo_agent_framework} illustrates \pipeline{EO-Gym}, a Gymnasium-style framework evaluating agents through stateful data-gathering rather than static inputs. This exploratory design is vital for EO evaluation, enabling agents to iteratively issue tool calls and gather evidence from new observations until resolving the query or reaching the rollout budget. The system operates over a local data lake comprising $660$k multimodal files with precise spatial, temporal, and sensor metadata. We construct this data lake using eight open source EO datasets \citep{lam2018xview,xia2018dota,li2020dior,sun2022fair1m,li2024sardet100k,christie2018fmow,gupta2019xbd,wang2025m4sar}. For example, we integrate \dataset{FAIR1M} \citep{sun2022fair1m} for geolocation, \dataset{xBD} \citep{gupta2019xbd} for disaster assessment, \dataset{fMoW} for temporal information, and \dataset{M4-SAR} \citep{wang2025m4sar} for cross-modal tasks.

\paragraph{Data gathering space.}
\pipeline{EO-Gym} enables agents to actively acquire evidence through an executable action space of $35$ EO-specialized tools. These tools return heterogeneous observations, including image crops, JSON bounding boxes, masked images, and spectral indices such as NDVI and NDWI \citep{rouse1974monitoring,gao1996ndwi}. We organize tool-mediated data access into three core paradigms: \textbf{Spatial Planning}, which supports panning and zooming beyond the initial crop; \textbf{Temporal Fetching}, which retrieves historical observations for a location; and \textbf{Cross-Modal Switching}, which uses a shared geospatial backbone to align optical and SAR views. All retrieval operations are implemented as local lookups, ensuring reproducible, low-latency interaction without dependence on live cloud platforms \citep{gorelick2017gee}. As an iterative engine, \pipeline{EO-Gym} operationalizes the interactive reasoning capabilities absent from static EO datasets.

\paragraph{Operation configuration.}
To support large-scale trajectory synthesis and evaluate agent behavior across diverse execution conditions, we vary three operational axes. First, the \textbf{tool response} is configured as either \textbf{Verified} or \textbf{Unverified}. To isolate planning quality, the Verified mode replaces non-deterministic optical object detection tools with ground-truth-backed simulator outputs. Instead of naively returning noisy open-source annotations, this mode generates bounding box candidates via optical detection models \cite{meta2025sam3}\cite{liu2023groundingdino}, which are then cross-validated against the ground truth using \model{GPT-4.1-mini} \cite{openai2026gpt41model} as a visual judge to select the most accurate observation. In contrast, the Unverified mode returns raw tool outputs to assess end-to-end robustness. For SAR and cross-modal tasks lacking such detectors, we directly utilize the ground truth, applying a Large Language Model (LLM) as semantic filter to realistically align with the agent's specific query. Second, \textbf{prompting} is designated as either \textbf{Simple} or \textbf{Detailed}. The Simple prompt solely instructs the agent to gather evidence via function calling, whereas the Detailed prompt provides high-level guidance on temporal, geospatial, and cross-modal access without revealing specific tool names or parameters. Third, the \textbf{tool schema list} is set to either \textbf{Skill} or \textbf{All}. The Skill mode restricts the action space by exposing only task-relevant tools, while the All mode exposes the complete $35$-tool schema to simulate an unconstrained EO workspace. Additional details are available in Appendix~\ref{operation-config-appendix}.

\section{Earth Observation Trajectory Dataset: \dataset{EO-Gym-Data}}

\begin{figure*}[t]
    \centering
    \includegraphics[width=\textwidth]{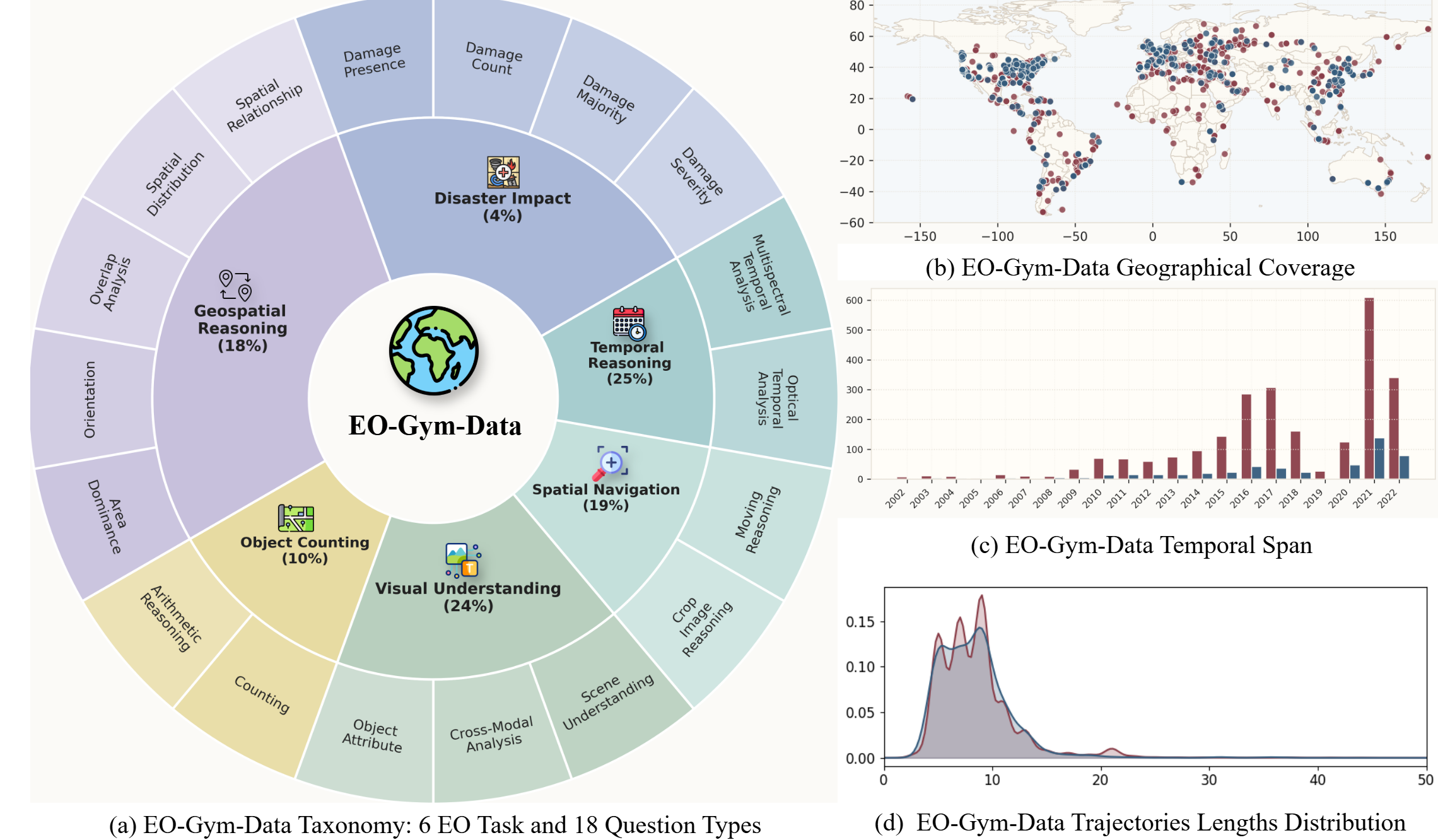}
    \caption{
    Overview of \dataset{EO-Gym-Data} statistics. 
    Panel (a) shows the taxonomy of six EO task categories and 18 question types. 
    Panel (b) shows the geographical distribution of the geolocated trajectory subset only, with red markers for training examples and blue markers for held-out test examples. 
    Panels (c) and (d) summarize the dataset's temporal spans and trajectory-length distribution, illustrating diversity in observation history and multi-step interaction depth.
    }
    \label{fig:dataset_stats}
\end{figure*}

\begin{figure*}[t]
    \centering
    \includegraphics[width=\textwidth]{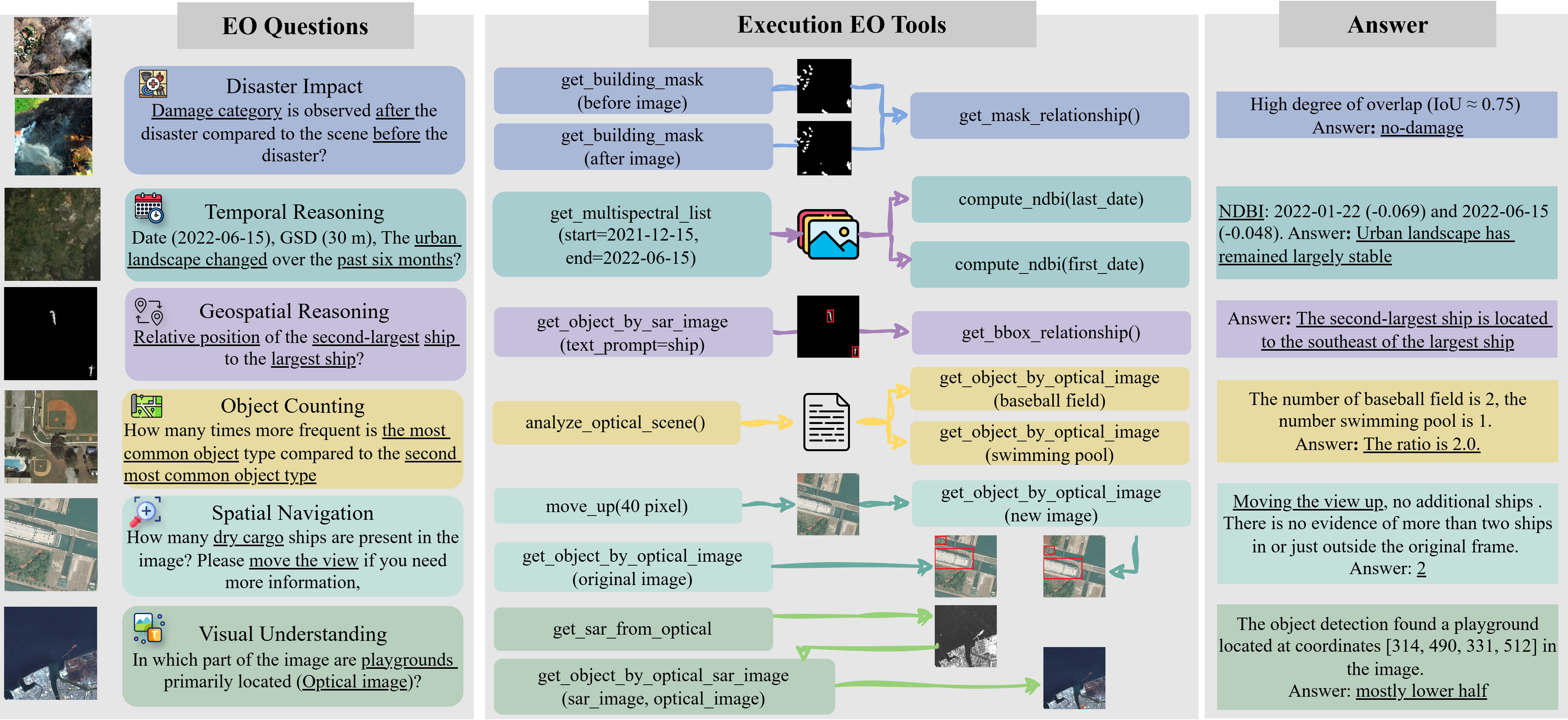}
    \caption{Qualitative examples from \dataset{EO-Gym-Data} across six EO task categories. Each example shows a complete interactive trajectory: an EO question (left), the sequential execution of EO tools to gather spatial, temporal, or cross-modal evidence (center), and the resulting verifiable answer (right).}
    \label{fig:eo_gym_data_example}
\end{figure*}

\begin{wraptable}{r}{0.6\textwidth}
\vspace{-15pt}
\footnotesize
\centering
\caption{Comparison with recent EO agent resources. \textit{DG} denotes whether the benchmark supports data-gathering execution tools for active evidence acquisition. \dataset{EO-Gym-Data} uniquely provides a training-scale VLM benchmark backed by massive indexed files with data-gathering tools.}
\label{tab:dataset-comparison}
\setlength{\tabcolsep}{3pt}
\begin{tabular}{l c c c}
\toprule
Dataset & VLM & DG & \# Tools / Traj. / Indexed Files \\
\midrule
ThinkGeo \citep{shabbir2025thinkgeo} & -- & -- & 14 / 486 / 0.36k \\
Earth-Bench \citep{feng2026earthagent} & \checkmark & \checkmark- & 104 / 248 / 13.7k \\
OpenEarth-Bench \citep{zhao2026openearthagent} & -- & -- & N/A / 596 / N/A \\
OpenEarthAgent \citep{shabbir2026openearthagent} & -- & -- & 24 / 15.7k / 12k \\
\midrule
\textbf{\dataset{EO-Gym-Data} (Ours)} & \textbf{\checkmark} & \textbf{\checkmark} & \textbf{35 / 9.1k / >660k} \\
\bottomrule
\end{tabular}
\vspace{-10pt}
\end{wraptable}

We construct \dataset{EO-Gym-Data} as a large-scale trajectory benchmark for training and evaluating interactive EO agents. The dataset contains $9{,}078$ trajectories and $34{,}604$ reasoning steps across six EO tasks and $18$ question types, with a held-out test split of $1{,}436$ manually verified trajectories and $7{,}642$ LLM-validated training trajectories. To prevent leakage, we partition training and test data by source imagery and question type, and mask filenames as hashes to avoid temporal, geolocation, or source-dataset shortcuts. Figure~\ref{fig:dataset_stats} summarizes the dataset scale, geolocated coverage, and temporal diversity, while Figure~\ref{fig:eo_gym_data_example} shows representative trajectories requiring stateful evidence gathering, such as historical-index comparison in \textit{Temporal Reasoning} and crop expansion in \textit{Spatial Navigation}. As shown in Table~\ref{tab:dataset-comparison}, \dataset{EO-Gym-Data} uniquely combines VLM support, training-scale trajectories, and a large data-gathering workspace indexing over $660$k true-color, multispectral, and SAR files with precise geospatial and temporal metadata. Details are provided in Appendix~\ref{data-overview-appendix}.

\paragraph{Candidate question-answer construction.}
We construct candidate items by sampling from the original dataset's ground truth to maximize label coverage. A shared generator maps these labels into one of 18 question types (Figure~\ref{fig:dataset_stats}a), each tied to a specific prompt that directs GPT-4.1 \citep{openai2026gpt41model} (temperature = 0.7) to produce an intermediate multiple-choice question (MCQ) and a reference answer. We use this MCQ format exclusively during trajectory synthesis to simplify generation and verification. For tasks lacking static targets, we create deferred-answer items and dynamically resolve their ground truth during pipeline execution using extracted physical quantities (e.g., NDVI).

\paragraph{Iterative guided synthesis.}
\label{sec:verified-mode}
We use an iterative synthesis pipeline to convert MCQ items into trajectories. During the \textbf{interaction} stage, \model{GPT-4.1} (temperature = $0.7$) acts as the tool-using generator. To maximize generation quality, the agent operates under the Verified mode, Detailed prompt, and Skill tools configuration, interacting with \pipeline{EO-Gym} for up to $50$ function calls. Following a generation attempt, the \textbf{verification} stage validates any synthesized trajectory reaching the predefined target answer by using \model{GPT-5.1} \citep{openai2025gpt51} (temperature = 0.3) to audit the trace for internal consistency, tool-use correctness, and evidence grounding. Specifically, if the auditor rejects a trajectory for exhibiting inconsistencies between the intermediate analysis and the final answer, it returns structured feedback to trigger a repair attempt by \model{GPT-4.1}. This feedback provides \textbf{guidance} on tool selection without leaking the target answer. We allow a maximum of three synthesis-review cycles.

\paragraph{Dataset enhance and review.}
From approximately 15k initial MCQ items, our pipeline successfully synthesizes valid traces for nearly 62\%. In the \textbf{enhance} stage, \model{gpt-oss-120b} \citep{openai2025gptoss120b} (temperature = 0.3) rewrites these traces into an explicit ReAct format \citep{yao2023react}. This standardizes rationales, sharpens action justifications, and converts the MCQ structure into an open-text format to minimize option-based guessing \citep{song2024agentbank,kang2025agentdistill}. We partition the data into two sets. For the test set, rigorous \textbf{human verification} refines the $1{,}650$ candidates into a final set of $1{,}436$ high-quality trajectories. Concurrently, for the training set, \textbf{automated validation} via \model{GPT-4.1} (temperature = 0.3), prompted with distilled human review criteria, strictly discards unsatisfactory trajectories, filtering $7{,}768$ candidates down to $7{,}642$.

%Our pipeline begins with approximately $15k$ candidate items and successfully synthesizes valid traces for nearly 62\% of these cases. Subsequently, \model{gpt-oss-120b} \citep{openai2025gptoss120b} (temperature = $0.3$) rewrites them into an explicit ReAct format \citep{yao2023react}. This \textbf{enhance} stage standardizes rationales, sharpens action justifications, and converts the MCQ format into a open-text format, reducing option-based guessing \citep{song2024agentbank,kang2025agentdistill}. We partition the resulting data into a test set for \textbf{human verification} and a training set for \textbf{automated validation} using \model{GPT-4.1} (temperature = $0.3$) prompted with distilled human review experience. By strictly discarding unsatisfactory trajectories rather than attempting repairs, the automated review filters the initial $7{,}768$ training candidates down to a final training set of $7{,}642$ trajectories. Concurrently, rigorous human verification refines the $1{,}650$ test candidates into a high-quality test set of $1{,}436$ trajectories.

\begin{table*}[t]
\centering
\footnotesize
\caption{Main benchmark results. \model{EO-Gym-4B} achieves the highest overall performance, leading in Pass@1, Pass@2, Pass@3, and Tool-any. Compared to its base model (\model{Qwen3-VL-4B-Instruct}), \model{EO-Gym-4B} yields substantial absolute improvements across all metrics, while simultaneously reducing unnecessary tool calls and formatting errors. \model{EO-Gym-4B} maintains a highly competitive, near-zero illegal error rate while vastly outperforming them in final task success.}
\label{tab:main-simple-prompt}
\resizebox{\textwidth}{!}{%
\begin{tabular}{lcccccc}
\toprule
Model & P@1 $\uparrow$ & P@2 $\uparrow$ & P@3 $\uparrow$ & Tool-call ratio $\downarrow$ & Illegal err. $\downarrow$ & Tool-any $\uparrow$ \\
\midrule
Qwen3-VL-4B-Instruct 
& 0.42 [0.39, 0.44] 
& 0.47 [0.44, 0.49] 
& 0.49 [0.46, 0.51] 
& 2.95 
& 0.040 
& 0.572 \\

Qwen3-VL-4B-Thinking 
& 0.36 [0.33, 0.38] 
& 0.45 [0.42, 0.48] 
& 0.49 [0.46, 0.51] 
& \textbf{1.71} 
& 0.006 
& 0.629 \\

Qwen3-VL-8B-Instruct 
& 0.45 [0.42, 0.47] 
& 0.47 [0.45, 0.50] 
& 0.49 [0.46, 0.52] 
& 2.69 
& 0.009 
& 0.512 \\

Qwen3-VL-8B-Thinking 
& 0.39 [0.36, 0.42] 
& 0.47 [0.45, 0.50] 
& 0.52 [0.50, 0.55] 
& 1.75 
& \textbf{0.002} 
& 0.619 \\

Qwen3-VL-32B-Instruct 
& 0.51 [0.49, 0.54] 
& 0.57 [0.54, 0.60] 
& 0.59 [0.57, 0.62] 
& 2.71 
& 0.009 
& 0.638 \\

GPT-4.1-mini 
& 0.56 [0.53, 0.58] 
& 0.61 [0.58, 0.63] 
& 0.64 [0.61, 0.66] 
& 3.37 
& 0.012 
& 0.697 \\

GPT-4.1 
& 0.57 [0.54, 0.59] 
& 0.61 [0.59, 0.64] 
& 0.64 [0.61, 0.66] 
& 2.97 
& 0.024 
& 0.664 \\

GPT-5.1 
& 0.53 [0.51, 0.56] 
& 0.59 [0.57, 0.62] 
& 0.63 [0.61, 0.66] 
& 2.25 
& 0.014 
& 0.368 \\

Gemini-2.5-Flash 
& 0.54 [0.51, 0.57] 
& 0.60 [0.57, 0.62] 
& 0.63 [0.60, 0.65] 
& 3.18 
& 0.003 
& 0.694 \\

Gemini-2.5-Pro 
& 0.54 [0.52, 0.57] 
& 0.63 [0.61, 0.65] 
& 0.67 [0.65, 0.69] 
& 3.02 
& 0.004 
& 0.617 \\
\midrule
\model{EO-Gym-4B} (Ours) 
& \textbf{0.65 [0.63, 0.68]} 
& \textbf{0.71 [0.69, 0.73]} 
& \textbf{0.74 [0.71, 0.76]} 
& 2.51 
& 0.003 
& \textbf{0.760} \\
\midrule
\model{EO-Gym-4B} (Ours) vs. Qwen3-VL-4B-Instruct 
& $+0.23$ 
& $+0.24$ 
& $+0.25$ 
& $-0.44$ 
& $-0.037$ 
& $+0.188$ \\
\bottomrule
\end{tabular}%
}
\end{table*}

\begin{table*}[t]
\centering
\footnotesize
\caption{The Table~\ref{tab:main-simple-prompt} Pass@3 results broken down by EO task. \model{EO-Gym-4B} achieves the strongest performance on Temporal, Disaster Impact, and Visual Understanding tasks.}
\label{tab:task-breakdown}
\resizebox{\textwidth}{!}{%
\begin{tabular}{lcccccc}
\toprule
Model & Temporal $\uparrow$ & Disaster Impact $\uparrow$ & Geospatial $\uparrow$ & Visual Underst. $\uparrow$ & Spatial Nav. $\uparrow$ & Object Counting $\uparrow$ \\
\midrule
Qwen3-VL-4B-Instruct & 0.29[0.24, 0.35] & 0.30 [0.18, 0.41] & 0.41 [0.34, 0.47] & 0.73[0.68, 0.78] & 0.60[0.54, 0.65] & 0.57 [0.47, 0.64] \\
Qwen3-VL-4B-Thinking & 0.09 [0.06, 0.13] & 0.23 [0.11, 0.34] & 0.47[0.41, 0.53] & 0.76[0.72, 0.81] & 0.75 [0.70, 0.80] & 0.46 [0.37, 0.54] \\
Qwen3-VL-8B-Instruct & 0.22 [0.18, 0.26] & 0.18 [0.09, 0.32] & 0.45[0.39, 0.52] & 0.71 [0.67, 0.76] & 0.64 [0.58, 0.69] & 0.61[0.52, 0.70] \\
Qwen3-VL-8B-Thinking & 0.10 [0.07, 0.13] & 0.25[0.14, 0.39] & 0.57 [0.51, 0.64] & 0.73 [0.69, 0.78] & 0.81[0.76, 0.85] & 0.61 [0.52, 0.71] \\
Qwen3-VL-32B-Instruct & 0.33[0.28, 0.38] & 0.50 [0.36, 0.64] & 0.53[0.47, 0.60] & 0.75[0.71, 0.80] & 0.76 [0.71, 0.81] & 0.66 [0.58, 0.74] \\
GPT-4.1-mini & 0.30 [0.26, 0.35] & 0.55[0.39, 0.68] & 0.64 [0.58, 0.70] & 0.79 [0.75, 0.83] & 0.83 [0.78, 0.87] & \textbf{0.80[0.73, 0.87]} \\
GPT-4.1 & 0.39[0.34, 0.44] & 0.59[0.43, 0.73] & 0.65 [0.59, 0.71] & 0.72[0.67, 0.77] & 0.80[0.75, 0.85] & 0.77 [0.70, 0.84] \\
GPT-5.1 & 0.50 [0.45, 0.54] & 0.57[0.43, 0.73] & \textbf{0.68 [0.62, 0.73]} & 0.60 [0.55, 0.65] & 0.78[0.73, 0.82] & 0.74 [0.67, 0.82] \\
Gemini-2.5-Flash & 0.34[0.29, 0.39] & 0.50 [0.36, 0.64] & 0.51 [0.44, 0.58] & 0.81[0.77, 0.85] & 0.81[0.77, 0.86] & 0.79 [0.71, 0.85] \\
Gemini-2.5-Pro & 0.42 [0.37, 0.47] & 0.52 [0.39, 0.68] & 0.60 [0.53, 0.66] & 0.81[0.77, 0.85] & \textbf{0.86[0.81, 0.90]} & 0.78 [0.70, 0.84] \\
\midrule
\model{EO-Gym-4B} (Ours) & \textbf{0.69 [0.65, 0.74]} & \textbf{0.68 [0.55, 0.82]} & 0.67 [0.61, 0.72] & \textbf{0.83[0.79, 0.87]} & 0.77 [0.72, 0.82] & 0.71 [0.63, 0.78] \\
\midrule
\model{EO-Gym-4B} (Ours) vs. Qwen3-VL-4B-Instruct 
& $+0.40$ & $+0.38$ & $+0.26$ & $+0.10$ & $+0.17$ & $+0.14$ \\
\bottomrule
\end{tabular}%
}
\end{table*}

\section{Model Fine-Tuning and Evaluation Protocol}
\label{sec:sft-and-evluation} 

We instantiate \model{EO-Gym-4B} by LoRA fine-tuning \model{Qwen3-VL-4B-Instruct} \citep{hu2022lora, bai2025qwen3vl} via SWIFT \citep{zhao2025swift} on \dataset{EO-Gym-Data} for three epochs (using the final checkpoint). To validate our dataset, we compare it against 10 native function-calling VLMs, strictly isolating planning from formatting mechanics. Open-weight baselines include the \model{Qwen3-VL} series (4B/8B Instruct and Thinking, 32B Instruct) \citep{bai2025qwen3vl}. Closed baselines comprise \model{GPT-4.1-mini}, \model{GPT-4.1}, \model{GPT-5.1}, \model{Gemini-2.5-Flash}, and \model{Gemini-2.5-Pro} \citep{openai2026gpt41model,openai2025gpt51,comanici2025gemini25}. Reasoning is configured with a 40,960 token limit for \model{Qwen3-VL-Thinking}, ``medium'' for \model{GPT-5.1}, and activated for \model{Gemini-2.5-Pro}. We exclude EO-domain VLMs (e.g., \model{EarthGPT}, \model{EarthDial} \citep{zhang2024earthgpt,soni2025earthdial}) because they lack native function calling. We exclude existing EO agent benchmarks (Table~\ref{tab:dataset-comparison}) because they are designed for text-only LLMs, are limited in scale, and lack the active data-gathering tools. All models are evaluated through their built-in tool schema. We conduct all training and open-weight inference on a single node equipped with four NVIDIA H200 GPUs.

\begin{wraptable}{r}{0.4\textwidth}
\vspace{-15pt}
\footnotesize
\centering
\caption{Zero function-call rates under \textbf{main evaluation setting}.}
\label{tab:zero-call-rates-compact}
\setlength{\tabcolsep}{4pt}
\begin{tabular}{l c}
\toprule
Model & Zero-Call Rate \\
\midrule
Qwen3-VL-4B-Instruct & 1.30\% \\
Qwen3-VL-4B-Thinking & 35.98\% \\
Qwen3-VL-8B-Instruct & 0.07\% \\
Qwen3-VL-8B-Thinking & 31.94\% \\
Qwen3-VL-32B-Instruct & 0.14\% \\
GPT-4.1-mini & 0.00\% \\
GPT-4.1 & 0.04\% \\
GPT-5.1 & 0.19\% \\
Gemini-2.5-Flash & 1.56\% \\
Gemini-2.5-Pro & 3.10\% \\
\midrule
\textbf{\model{EO-Gym-4B} (Ours)} & \textbf{0.02\%} \\
\bottomrule
\end{tabular}
\vspace{-10pt}
\end{wraptable}

\paragraph{Evaluation protocol.}

We evaluate all models on the full held-out test set of $1{,}436$ examples, with at most $15$ function calls per rollout. We employ \model{GPT-4.1-mini} as a fixed semantic judge to evaluate the binary correctness of open-ended text responses against the ground truth. On a balanced human-annotated subset of $600$ outputs, the judge achieves $0.923$ observed agreement and Cohen's $\kappa=0.846$ \citep{cohen1960coefficient}, supporting its reliability for large-scale reporting. We define zero-function-call trajectories as failures to evaluate data-gathering capacity. As summarized in Table ~\ref{tab:zero-call-rates-compact}, zero-call failures are rare for EO-GYM-4B, GPT models, and non-thinking Qwen variants, but frequent for Qwen-Thinking variants. Unless otherwise specified, our \textbf{main evaluation setting} employs \textbf{Verified mode}, \textbf{Simple prompt}, and the \textbf{Skill} tools setting.

\paragraph{Evaluation metrics.} 
%Our primary metric is \textbf{Pass@$k$} \citep{chen2021codex}, which measures the probability that at least one of $k$ independently sampled trajectories reaches a semantically correct final answer. We report point estimates for Pass@$k$ with the $\pm$ half-width of the 95\% bootstrap confidence interval. We complement this with execution diagnostics. \textbf{Illegal err.} measures the average frequency of parameter content errors during function calls, such as passing an empty string as input. \textbf{Tool-call ratio} reports the average ratio between the model's function-call count and the ground-truth call count on successful trajectories. Finally, \textbf{Tool-any} measures the unordered overlap between invoked tools and the reference tool set on successful trajectories. We intentionally avoid order-sensitive alignment metrics because many EO workflows contain commutative subroutines.

Our primary metric is \textbf{Pass@$k$} \citep{chen2021codex}, which calculates the probability of finding a successful solution within $k$ attempts. We report point estimates for Pass@$k$ with the $\pm$ half-width of the 95\% bootstrap confidence interval. For each evaluation example, we generate $n$ total trajectories, of which $c$ are successful. The unbiased estimator used across the dataset is:
$$\mathrm{Pass@}k = \mathbb{E}\left[1 - \frac{\binom{n-c}{k}}{\binom{n}{k}}\right]$$

We complement final-answer metrics with execution diagnostics computed over \textit{consumed trajectories}, defined as the attempts evaluated sequentially until the first correct final answer, or all available attempts if no correct answer is found. \textbf{Illegal err.} measures the average per-question rate of consumed function calls whose observations are missing, empty, or explicitly indicate execution errors. \textbf{Tool-call ratio} reports the ratio between the number of consumed model function calls and the reference function-call count, summarized across questions. Finally, \textbf{Tool-any} measures the average consumed-trajectory rate of binary any-order tool coverage, where a trajectory is counted as matched if its invoked-tool multiset contains all reference tools.

\section{Experiment}
\label{sec:ablations}

\paragraph{Main benchmark results.}
Table~\ref{tab:main-simple-prompt} presents the headline result. Under the Verified mode Simple prompt protocol, \model{EO-Gym-4B} is the strongest model overall, reaching \textbf{0.65} Pass@1, \textbf{0.71} Pass@2, and \textbf{0.74} Pass@3. Relative to its untuned \model{Qwen3-VL-4B-Instruct} base, this yields absolute gains of +0.23, +0.24, and +0.25. The improvement is not marginal: \model{EO-Gym-4B}'s Pass@3 confidence interval, \textbf{[0.71, 0.76]}, lies entirely above that of the strongest baseline, \model{Gemini-2.5-Pro}, at 0.67 [0.65, 0.69]. The gain also extends beyond final-answer accuracy. Among successful tool-using trajectories, EO-GYM-4B achieves the highest Tool-any score of 0.760, reduces the tool-call ratio from 2.95 to 2.51 relative to its untuned base model, and maintains a low illegal-call rate of 0.003. These results demonstrate that EO-specific fine-tuning improves both benchmark success and the fidelity of executable behavior.

\paragraph{Task-level performance breakdown.}
To understand the source of this overall gain, we next analyze performance across the six primary EO task categories. Table~\ref{tab:task-breakdown} details the Pass@3 results. \model{EO-Gym-4B} improves most on tasks requiring active data gathering. In \textit{Temporal Reasoning} and \textbf{Disaster Impact}, it reaches 0.69 and 0.68, far exceeding the 0.29 and 0.30 scores of its untuned base model. It also remains highly competitive in \textit{Geospatial Reasoning} at 0.67, closely following \model{GPT-5.1} at 0.68. Conversely, closed models remain highly competitive on perception-heavy tasks. \model{Gemini-2.5-Pro} leads \textit{Spatial Navigation} at 0.86, and \model{GPT-4.1-mini} leads \textit{Object Counting} at 0.80. This performance distribution confirms that EO-specific fine-tuning successfully strengthens sequential evidence gathering across space, time, and modality. 

\paragraph{Stability under repeated sampling.}
We also investigate whether strong Pass@$k$ values arise from sampling luck rather than stable procedural competence. We increase the sampling budget to $k=10$ and report results in Table~\ref{tab:pass10-metrics}. We also report \textbf{Self-Consistency}, which evaluates the correctness of the majority-voted answer \citep{wang2022selfconsistency}. \model{EO-Gym-4B} reaches 0.80 Pass@10. It also achieves the highest Self-Consistency score of 0.67. These results show that \model{EO-Gym-4B} produces denser correct generations and more stable majority behavior under repeated sampling.

\paragraph{Robustness to prompt specification.}
A critical question is whether stronger prompt specification eliminates the need for EO-specific fine-tuning. Table~\ref{tab:main-complex-prompt} shows that the \textbf{Detailed prompt} setting makes high-capacity generalist models more competitive, but does not remove the benefit of EO-specific supervision. \model{EO-Gym-4B} achieves the highest Pass@1 score of 0.60 and the highest Tool-any score of 0.72. \model{GPT-5.1} slightly leads on Pass@2 and Pass@3, reaching 0.66 and 0.70, respectively, compared with 0.66 and 0.69 for \model{EO-Gym-4B}. Relative to its base model, \model{Qwen3-VL-4B-Instruct}, \model{EO-Gym-4B} improves Pass@1, Pass@2, Pass@3, and Tool-any by +0.16, +0.16, +0.15, and +0.14, respectively. These results indicate that detailed prompting narrows the gap, but EO-specific fine-tuning still yields stronger tool-grounded behavior.

\begin{table}[htbp]
\centering
\scriptsize
\caption{\textbf{Repeated sampling} with $k=10$. \model{EO-Gym-4B} consistently outperforms baseline models}
\label{tab:pass10-metrics}
\resizebox{\textwidth}{!}{%
\begin{tabular}{l c c c c}
\toprule
Model & Pass@5 $\uparrow$ & Pass@7 $\uparrow$ & Pass@10 $\uparrow$ & Self-Cons. $\uparrow$ \\
\midrule
Qwen3-VL-4B-Instruct 
& 0.52 [0.49, 0.54] 
& 0.54 [0.51, 0.57] 
& 0.55 [0.52, 0.58] 
& 0.42 \\

Qwen3-VL-8B-Instruct 
& 0.51 [0.49, 0.54] 
& 0.52 [0.49, 0.55] 
& 0.53 [0.51, 0.56] 
& 0.45 \\

\model{EO-Gym-4B} (Ours) 
& \textbf{0.78 [0.76, 0.80]} 
& \textbf{0.79 [0.77, 0.81]} 
& \textbf{0.80 [0.78, 0.83]} 
& \textbf{0.67} \\
\bottomrule
\end{tabular}%
}
\end{table}

\begin{table*}[t]
\centering
\footnotesize
\caption{\textbf{Prompt specification} results under the \textbf{Detailed prompt} configuration. \model{EO-Gym-4B} achieves the best Pass@1 and Tool-any, while remaining competitive with the strongest generalist model on Pass@2 and Pass@3.}
\label{tab:main-complex-prompt}
\resizebox{\textwidth}{!}{%
\begin{tabular}{lcccccc}
\toprule
Model & P@1 $\uparrow$ & P@2 $\uparrow$ & P@3 $\uparrow$ & Tool-call ratio $\downarrow$ & Illegal err. $\downarrow$ & Tool-any $\uparrow$ \\
\midrule
Qwen3-VL-4B-Instruct & 0.44 [0.42, 0.47] & 0.50 [0.48, 0.53] & 0.54 [0.52, 0.57] & 2.33 & 0.06 & 0.58 \\
Qwen3-VL-4B-Thinking & 0.35 [0.32, 0.37] & 0.44 [0.41, 0.46] & 0.48 [0.46, 0.51] & \textbf{1.57} & 0.01 & 0.61 \\
Qwen3-VL-8B-Instruct & 0.49 [0.46, 0.52] & 0.56 [0.53, 0.58] & 0.58 [0.56, 0.61] & 2.62 & 0.04 & 0.57 \\
Qwen3-VL-8B-Thinking & 0.38 [0.35, 0.40] & 0.48 [0.45, 0.50] & 0.53 [0.50, 0.55] & 1.62 & 0.01 & 0.60 \\
Qwen3-VL-32B-Instruct & 0.48 [0.45, 0.51] & 0.55 [0.52, 0.57] & 0.59 [0.57, 0.62] & 2.73 & 0.04 & 0.62 \\
GPT-4.1-mini & 0.55 [0.53, 0.58] & 0.63 [0.60, 0.65] & 0.66 [0.64, 0.69] & 3.58 & 0.03 & 0.68 \\
GPT-4.1 & 0.53 [0.51, 0.56] & 0.60 [0.57, 0.62] & 0.62 [0.60, 0.65] & 2.83 & 0.04 & 0.64 \\
GPT-5.1 & 0.59 [0.56, 0.61] & \textbf{0.66 [0.64, 0.69]} & \textbf{0.70 [0.68, 0.73]} & 2.33 & 0.01 & 0.45 \\
Gemini-2.5-Flash & 0.53 [0.51, 0.56] & 0.60 [0.58, 0.63] & 0.64 [0.62, 0.67] & 3.06 & 0.01 & 0.67 \\
Gemini-2.5-Pro & 0.50 [0.47, 0.52] & 0.61 [0.58, 0.63] & 0.64 [0.62, 0.67] & 2.79 & \textbf{0.00} & 0.57 \\
\midrule
\model{EO-Gym-4B} (Ours) & \textbf{0.60 [0.57, 0.62]} & 0.66 [0.64, 0.68] & 0.69 [0.67, 0.71] & 2.50 & 0.02 & \textbf{0.72} \\
\bottomrule
\end{tabular}%
}
\end{table*}

\begin{table*}[htbp]
\centering
\footnotesize
\caption{\textbf{RQ1: End-to-end evaluation} under the Unverified mode, All tools, and Simple prompt. This study tests whether the model survives raw tool outputs and the full execution space.}
\label{tab:main-unverifed-prompt-comparison}
\resizebox{\textwidth}{!}{%
\begin{tabular}{l c c c c c c}
\toprule
Model & P@1 $\uparrow$ & P@2 $\uparrow$ & P@3 $\uparrow$ & Tool-call ratio $\downarrow$ & Illegal err. $\downarrow$ & Tool-any $\uparrow$ \\
\midrule
Qwen3-VL-4B-Instruct & 0.39 [0.37, 0.42] & 0.45 [0.42, 0.47] & 0.48 [0.45, 0.51] & 2.51 & 0.019 & 0.429 \\
Qwen3-VL-8B-Instruct & 0.47 [0.44, 0.49] & 0.52[0.49, 0.54] & 0.54[0.51, 0.56] & 2.46 & 0.008 & 0.423 \\
Qwen3-VL-32B-Instruct & 0.53 [0.51, 0.56] & 0.58 [0.55, 0.60] & 0.61[0.59, 0.64] & 2.47 & 0.005 & 0.409 \\
GPT-4.1-mini & 0.53[0.51, 0.56] & 0.61[0.59, 0.64] & 0.65 [0.63, 0.67] & 2.75 & \textbf{0.003} & 0.429 \\
\model{EO-Gym-4B} (Ours) & \textbf{0.61[0.58, 0.63]} & \textbf{0.66 [0.63, 0.68]} & \textbf{0.69[0.66, 0.71]} & \textbf{2.23} & 0.005 & \textbf{0.456} \\
\bottomrule
\end{tabular}%
}
\end{table*}

\begin{table*}[t]
\centering
\small
\caption{\textbf{RQ2: Tool \& transfer evaluation}. The \textit{Base} model is the untuned \model{Qwen3-VL-4B-Instruct}. The \textit{SFT} column represents three separate models, each fine-tuned on a training set where the target tool group was withheld.}
\label{tab:transfer-ablation-pivoted-simple}
\resizebox{\textwidth}{!}{%
\begin{tabular}{l ccc ccc ccc}
\toprule
& \multicolumn{3}{c}{Temporal Fetching (TF)}
& \multicolumn{3}{c}{Spatial Planning (SP)}
& \multicolumn{3}{c}{Cross-modal Switching (CMS)} \\
& \multicolumn{3}{c}{\scriptsize \textit{(764 train, 265 test)}}
& \multicolumn{3}{c}{\scriptsize \textit{(635 train, 114 test)}}
& \multicolumn{3}{c}{\scriptsize \textit{(384 train, 104 test)}} \\
\cmidrule(lr){2-4} \cmidrule(lr){5-7} \cmidrule(lr){8-10}
Metric
& Base & SFT & $\Delta$
& Base & SFT & $\Delta$
& Base & SFT & $\Delta$ \\
\midrule
Pass@1
& 0.057 & 0.109 & +0.052
& 0.412 & 0.518 & +0.105
& 0.731 & 0.712 & -0.019 \\

Pass@2
& 0.083 & 0.162 & +0.079
& 0.456 & 0.640 & +0.184
& 0.788 & 0.827 & +0.038 \\

Pass@3
& 0.087 & 0.211 & +0.125
& 0.465 & 0.693 & +0.228
& 0.798 & 0.856 & +0.058 \\
\bottomrule
\end{tabular}%
}
\end{table*}

To validate \model{EO-Gym-4B}, we investigate three research questions (RQs) testing end-to-end robustness versus tool memorization, restricting these analyses to a representative model subset to manage costs.

\paragraph{RQ1: End-to-end evaluation.}
The verified mode removes detector noise and isolates planning quality. We ask whether this advantage survives when models consume raw tool outputs. To rigorously test this end-to-end robustness, we evaluate models using the All tools and Unverified configuration. Table~\ref{tab:main-unverifed-prompt-comparison} shows that the advantage persists. \model{EO-Gym-4B} achieves the strongest Pass@1, Pass@2, and Pass@3, reaching \textbf{0.61}, \textbf{0.66}, and \textbf{0.69}, respectively. Final-answer accuracy changes unevenly under the Unverified and All-tools setting, with some Qwen models showing slight gains. However, Tool-any drops for every evaluated model, indicating a consistent degradation in tool-use fidelity.

%\paragraph{RQ2: Disjoint-tool transfer.}
%We evaluate whether \model{EO-Gym-4B} can transfer operational knowledge beyond the exact tool. We construct three trajectory-level transfer splits and fine-tune a separate model for each setting using the configuration in Section~\ref{sec:sft-and-evluation}. \textit{Temporal Fetching} provides a clean modality-level tool holdout between optical and multispectral temporal access. \textit{Cross-modal Switching} evaluates directional bridge transfer by training on optical-to-SAR access and testing on the reverse direction, while retaining shared support tools. \textit{Spatial Planning} evaluates question-type transfer from relational reasoning to area and distribution analysis under substantial reuse of foundational navigation and detection tools. We do not isolate basic image-navigation tools, since they appear throughout long-horizon trajectories and removing them would make the resulting splits too small for reliable evaluation. Across the three settings, \model{EO-Gym-4B} improves over the base \model{Qwen3-VL-4B-Instruct} model except for a small Pass@1 decrease in Cross-modal Switching. The largest gain occurs in Spatial Planning, where Pass@3 improves by $+0.228$. These results indicate that EO-specific fine-tuning supports targeted transfer beyond memorized tool strings or single fixed workflows. We defer details to Appendix~\ref{sec:tool_group_splitting}.

\paragraph{RQ2: Tool \& transfer evaluation}
We evaluate whether \model{EO-Gym-4B} transfers operational knowledge beyond the exact tools and workflows seen during fine-tuning. We construct three trajectory-level transfer splits and fine-tune a separate model for each using the configuration in Section~\ref{sec:sft-and-evluation}. \textit{Temporal Fetching} provides a clean modality-level tool holdout between optical and multispectral temporal access. \textit{Cross-modal Switching} evaluates directional bridge transfer by training on optical-to-SAR access and testing on the reverse direction, while retaining shared support tools. \textit{Spatial Planning} evaluates question-type transfer from relational reasoning to area and distribution analysis under substantial reuse of foundational navigation and detection tools. We do not isolate basic image-navigation tools because they appear throughout long-horizon trajectories, making strict removal impractical. Across all settings, \model{EO-Gym-4B} improves over the base \model{Qwen3-VL-4B-Instruct} except for a small Pass@1 drop in Cross-modal Switching. These results suggest that EO-specific fine-tuning supports targeted transfer beyond memorized tool strings.

\begin{table}[htbp]
\centering
\scriptsize
\caption{\textbf{RQ3: Tool renaming} results. Renaming all exposed tools at inference time leaves performance and tool-use behavior essentially unchanged, indicating that \model{EO-Gym-4B} relies on functional semantics rather than memorized tool strings. }
\label{tab:tool-rename-simple}
\resizebox{\columnwidth}{!}{%
\begin{tabular}{lcccccc}
\toprule
Tool & Pass@1 $\uparrow$ & Pass@2 $\uparrow$ & Pass@3 $\uparrow$ & Tool-call ratio $\downarrow$ & Illegal err. $\downarrow$ & Tool-any $\uparrow$ \\
\midrule
Original 
& 0.65 [0.63, 0.68] 
& 0.71 [0.69, 0.73] 
& 0.74 [0.71, 0.76] 
& 2.51 
& 0.003 
& 0.760 \\

Rename   
& 0.66 [0.64, 0.69]
& 0.72 [0.70, 0.74]
& 0.75 [0.72, 0.77]
& 2.50
& 0.003
& 0.760 \\
\bottomrule
\end{tabular}%
}
\end{table}

\paragraph{RQ3: Tool renaming.}
We investigate whether the model's performance gain is an artifact of memorizing via tool names. Table~\ref{tab:tool-rename-simple} shows that performance remains essentially unchanged when all exposed tools are renamed at inference time. The renamed configurations yield nearly identical Pass@$k$ scores and execution metrics compared to the original names. This strict invariance is difficult to reconcile with string-level memorization. Instead, it suggests that \model{EO-Gym-4B} has learned a robust functional mapping from task requirements to tool affordances.

%Taken together, these diagnostics strengthen the interpretation of the main benchmark result. \model{EO-Gym-4B}'s advantage is not well explained by a single prompt template, lucky repeated sampling, or superficial memorization of tool identifiers. Rather, the evidence consistently points to a real gain in EO-specific interactive competence.

\section{Conclusion}
We presented \pipeline{EO-Gym}, a controlled executable environment that operationalizes Earth observation as an interactive evidence-acquisition problem. By moving beyond static input benchmarks, \pipeline{EO-Gym} requires agents to actively navigate space, time, and modality to resolve uncertainty. To support this paradigm, we introduced \dataset{EO-Gym-Data}, a massive trajectory benchmark grounded in over $660$k multimodal indexed files. Our extensive evaluation reveals that while leading general-purpose VLMs struggle with these dynamic workflows, EO-specific fine-tuning systematically closes the gap. The resulting \model{EO-Gym-4B} baseline achieves strong performance and demonstrates targeted transfer across tool-families. Ultimately, this framework provides a rigorous, reproducible foundation for developing autonomous agents capable of planetary-scale geospatial reasoning.

\section{Limitations and Future Work}
Despite these advances, our framework has several limitations. First, we do not explicitly test knowledge transfer across the three core data-access paradigms. Because these paradigms are tightly coupled within long-horizon trajectories, enforcing strict isolation would leave fewer than 10 examples per split, which is insufficient for meaningful fine-tuning or benchmarking. Second, EO-Gym currently uses a predefined tools, although the framework can be extended by registering new Python functions and adding their descriptions to the environment schema. Future versions could further support open-ended code generation or dynamic tool creation for unconstrained operational queries. Finally, while our fine-tuning focuses on a 4B model to isolate the effect of specialized supervision, scaling training to larger foundation models may yield stronger interactive reasoning.

\section{Ethical and Licensing Considerations}
To support the research community, we have open-sourced the \dataset{EO-Gym} dataset and \model{EO-Gym-4B}. \dataset{EO-Gym} is built from publicly available datasets, and follows the licenses and access terms of the original providers. The satellite and aerial imagery used in our benchmark does not support direct identification of individuals, reducing privacy risks. However, EO data may still expose sensitive contextual information when combined with geolocation, temporal metadata, or external sources, especially for sensitive sites, disaster-affected regions, remote communities, Indigenous lands, or conflict-affected areas. The benchmark may inherit geographic, sensor, seasonal, and label-distribution biases from its sources, and EO tools can be dual-use. We release \dataset{EO-Gym} for reproducible research in environmental monitoring, climate science, and disaster response, and do not endorse individual tracking. Generated trajectories and model outputs may still contain residual errors, and multimodal agent training has computational and environmental costs.

\appendix

\section{Controlled Executable Earth Observation Environment: \pipeline{EO-Gym}}

\subsection{Metadata-Drive Data Lake}
\label{data-source}

The \pipeline{EO-Gym} environment operationalizes EO reasoning as an interactive evidence-acquisition process across space, time, and modality. To construct a comprehensive data gathering space, the environment leverages eight open-source dataset imagery collections alongside their corresponding ground truth and metadata. Furthermore, we augment this corpus by retrieving Landsat and Sentinel-2 multispectral imagery via Google Earth Engine \cite{gorelick2017gee}. Table \ref{tab:eogym-datasets} summarizes these sources.

For the \dataset{Functional Map of the World (fMoW)} dataset \cite{christie2018fmow}, the environment consumes local true color (RGB) frame sequences accompanied by frame-level JSON metadata and sequence mapping archives. The ground truth links bounding-box identifiers to category names while filtering out false detections. The retained metadata includes sequence identifiers, frame indices, modalities, timestamps, ground sample distance, image dimensions, and bounding-box coverage statistics. These temporal sequences support visual question answering steps by comparing changes across frames using temporal order and object labels.

The \dataset{xBD} dataset \cite{gupta2019xbd} utilizes pre-disaster and post-disaster GeoTIFF tiles converted into RGB images. The ground truth consists of label polygons converted into pixel-space building footprints. The metadata carries forward source paths, image dimensions, feature unique identifiers, feature types, and damage subtypes. The paired before and after imagery enables damage presence, damage count, majority damage, and severity reasoning.

The \dataset{FAIR1M} dataset \cite{sun2022fair1m} employs very high-resolution (VHR) optical tiles exported alongside GeoJSON area-of-interest anchors. The ground truth provides pixel-space polygons and WGS84 geometries for fine-grained object recognition. The metadata preserves image coordinate reference systems, ground sample distance, capture times, and image centers. The area-of-interest geometries and capture times serve as anchors to query Google Earth Engine for Sentinel-2 and Landsat images within a surrounding time window. By leveraging the fine-grained object labels from \dataset{FAIR1M}, we can infer broader scene classifications, such as identifying urban or industrial zones based on high vehicle density. This approach facilitates complex multispectral temporal questions that combine paired sensor data with the inferred scene context derived from the original object annotations.

We integrate the \dataset{DOTA} \cite{xia2018dota}, \dataset{xView} \cite{lam2018xview}, and \dataset{DIOR} \cite{li2020dior} datasets to provide diverse optical imagery. The ground truth annotations are parsed into category names, bounding boxes, object areas, and polygon rings. The preserved metadata includes dataset origins, splits, image identifiers, dimensions, and bounding box formats. These datasets are sampled based on object-label coverage to support counting, spatial reasoning, overlap analysis, attribute identification, and image cut reasoning questions.

The \dataset{M4-SAR} dataset \cite{wang2025m4sar} incorporates paired optical and SAR imagery. The ground truth maps numeric classes to specific infrastructure and maritime categories such as bridges, harbors, oil tanks, playgrounds, airports, and wind turbines. The metadata tracks modality-specific image paths, class labels, and polygon boxes scaled to pixel coordinates. The cross-modal questions rely on these paired modality paths and shared object labels.

The \dataset{SARDet-100K} dataset \cite{li2024sardet100k} consumes standardized SAR imagery patches. The ground truth is derived from standard object detection annotations. The metadata retains image identifiers, category names, and bounding boxes. These records are balanced by object-label coverage to support modality-specific object counting, spatial distribution analysis, and arithmetic reasoning.

\begin{table*}[t]
\centering
\small
\caption{Open-source imagery sources used to construct \pipeline{EO-Gym}.}
\label{tab:eogym-datasets}
\begin{tabular}{p{2cm}p{3.2cm}p{5.0cm}p{2cm}}
\toprule
Dataset & Modality used in EO-Gym & Reported imagery source and platform & Ground Sample Distance (GSD) \\
\midrule
xBD \cite{gupta2019xbd} & Optical, temporal pre and post pairs &
True color pre and post disaster satellite imagery with building polygons, damage labels, and satellite metadata &
$\le$ 0.8 m GSD \\
\midrule
xView \cite{lam2018xview} & VHR optical &
WorldView-3 \cite{digitalglobe2014worldview3} satellite imagery &
0.3 m GSD \\
\midrule
DIOR \cite{li2020dior} & Optical &
Large-scale optical remote-sensing imagery with diverse imaging conditions and resolutions &
0.5 to 30 m GSD \\
\midrule
FAIR1M \cite{sun2022fair1m} & VHR optical and geolocation metadata &
High-resolution imagery from multiple platforms geographic information provided with each image &
0.3 to 0.8 m GSD \\
\midrule
fMoW \cite{christie2018fmow} & Optical, temporal metadata, and multispectral &
4-band imagery from QuickBird-2 \cite{toutin2002quickbird} and GeoEye-1 \cite{madden2009geoeye1} 8-band imagery from WorldView-2 and WorldView-3 \cite{digitalglobe2014worldview3} &
0.3 to 0.5 m GSD \\
\midrule
DOTA \cite{xia2018dota} & Optical and aerial &
Google Earth, GF-2 \cite{huang2018gf2}, JL-1 \cite{wmo2026jilin1}, and CycloMedia \cite{cyclomedia2026aerial} imagery &
0.1 to 4.5 m GSD \\
\midrule
SARDet-100K \cite{li2024sardet100k} & SAR &
Standardized aggregation of 10 SAR detection datasets &
0.5 to 25 m GSD \\
\midrule
M4-SAR \cite{wang2025m4sar} & Paired optical and SAR &
Sentinel-1 \cite{torres2012sentinel1} SAR paired with Sentinel-2 \cite{drusch2012sentinel} optical imagery &
10 to 60 m GSD \\
\midrule
GEE multispectral \cite{gorelick2017gee} & Moderate-resolution multispectral time series &
Landsat 8 and 9 \cite{wulder2019landsatstatus} and Sentinel-2 \cite{drusch2012sentinel} files retrieved from Google Earth Engine using FAIR1M geolocations &
10 to 60 m GSD \\
\bottomrule
\end{tabular}
\end{table*}

\subsection{Data Gathering Space}
\label{eo-tool-catalog}

The \pipeline{EO-Gym} environment exposes a comprehensive suite of $35$ specialized tools designed to support interactive EO reasoning. To systematically categorize their functionalities, we partition the tool space into two primary tool groups: data gathering tools and data analysis alongside support tools. 

The \textit{data gathering tools}, detailed in Table \ref{tab:data-gathering-tools}, are responsible for actively acquiring new evidence from the environment. These tools enable spatial planning through zooming and panning, temporal fetching to retrieve historical observations, and cross-modal switching to transition between aligned optical and SAR imagery. 

The \textit{data analysis and support tools}, detailed in Table \ref{tab:data-analysis-tools}, operate on the already selected images, masks, bounding boxes, or multispectral band folders. This group encompasses object detection, area masking, spectral index analysis, geospatial relationship computation, and semantic interpretation.

\begin{table*}[htbp]
\centering
\small
\caption{Data gathering tools in \pipeline{EO-Gym}. These tools actively acquire new spatial, temporal, or cross-modal evidence from the environment.}
\label{tab:data-gathering-tools}
\begin{tabular}{p{2.0cm}p{4.5cm}p{6.5cm}}
\toprule
Data Gathering Method & Tool Name & Main Role and Description \\
\midrule
\multirow{7}{2.0cm}{Spatial Planning} 
& \texttt{crop\_multispectral\_image} & AOI cropping and multispectral preparation. Crops selected multispectral band folders to a normalized AOI and saves NetCDF outputs for later index or mask analysis. \\
& \texttt{crop\_optical\_or\_sar\_image} & AOI cropping and image preparation. Crops an optical or SAR image to a normalized AOI. This serves as the zoom-in equivalent for focused inspection. \\
& \texttt{move\_down\_optical\_image} & Pan and navigation. Shifts the current crop downward while preserving the crop size. \\
& \texttt{move\_left\_optical\_image} & Pan and navigation. Shifts the current crop left while preserving the crop size. \\
& \texttt{move\_right\_optical\_image} & Pan and navigation. Shifts the current crop right while preserving the crop size. \\
& \texttt{move\_up\_optical\_image} & Pan and navigation. Shifts the current crop upward while preserving the crop size. \\
& \texttt{zoom\_out\_optical\_image} & Zoom and navigation. Expands the visible area around a cropped VHR image. \\
\midrule
\multirow{6}{2.0cm}{Temporal Fetching} 
& \texttt{get\_multispectral\_list} & Temporal and multispectral retrieval. Lists available Sentinel and Landsat multispectral observations for a FAIR1M location. \\
& \texttt{get\_next\_multispectral} & Temporal and multispectral retrieval. Retrieves the next multispectral folder after a reference date. \\
& \texttt{get\_next\_optical\_image} & Temporal and optical retrieval. Retrieves the closest later RGB image for the same optical sequence and location. \\
& \texttt{get\_optical\_image\_list} & Temporal and optical retrieval. Lists co-located RGB frames ordered by timestamp or frame index. \\
& \texttt{get\_previous\_multispectral} & Temporal and multispectral retrieval. Retrieves the most recent multispectral folder before a reference date. \\
& \texttt{get\_previous\_optical\_image} & Temporal and optical retrieval. Retrieves the closest earlier RGB image for the same optical sequence and location. \\
\midrule
\multirow{2}{2.0cm}{Cross-Modal Switching} 
& \texttt{get\_optical\_from\_sar} & Cross-modal retrieval. Returns the aligned optical companion for a SAR input. \\
& \texttt{get\_sar\_from\_optical} & Cross-modal retrieval. Returns the aligned SAR companion for an optical input. \\
\bottomrule
\end{tabular}
\end{table*}

\begin{table*}[htbp]
\centering
\small
\caption{Data analysis and support tools in \pipeline{EO-Gym}. These tools process acquired observations to extract objects, masks, spectral indices, and semantic relationships.}
\label{tab:data-analysis-tools}
\begin{tabular}{p{2cm}p{6.5cm}p{4.5cm}}
\toprule
Analysis Group & Tool Name & Main Role and Description \\
\midrule
\multirow{2}{2.0cm}{Semantic understanding} 
& \texttt{analyze\_optical\_scene} & Classifies or summarizes the scene from an optical image. It is not intended for counting or detection. \\
& \texttt{describe\_optical\_object} & Uses vision-language reasoning to describe visible attributes such as color or appearance. \\
\midrule
\multirow{3}{2.0cm}{Object Detection} 
& \texttt{get\_object\_bbox\_by\_optical\_image} & Detects prompt-specified objects in optical imagery and returns bounding boxes. \\
& \texttt{get\_object\_bbox\_by\_sar\_image} & Detects or filters prompt-specified objects in SAR imagery. This is commonly used after switching from optical to SAR. \\
& \texttt{get\_object\_bbox\_by\_optical\_sar\_image} & Uses an aligned optical and SAR pair to select or verify object boxes. \\
\midrule
\multirow{6}{2.0cm}{Area Masking} 
& \texttt{get\_building\_mask\_by\_optical\_image} & Extracts building masks from optical images. \\
& \texttt{get\_road\_mask\_by\_optical\_image} & Extracts road masks from optical images. \\
& \texttt{get\_object\_mask\_by\_optical\_image} & Segments prompt-specified objects in optical imagery and returns binary masks. \\
& \texttt{compute\_urban\_mask\_by\_multispectral} & Converts multispectral bands into an urban and built-up mask using NDBI-style thresholds. \\
& \texttt{compute\_vegetation\_mask\_by\_multispectral} & Converts multispectral bands into a vegetation mask using NDVI-style thresholds. \\
& \texttt{compute\_water\_mask\_by\_multispectral} & Converts multispectral bands into a water mask using NDWI-style thresholds. \\
\midrule
\multirow{5}{2.0cm}{Spectral Analysis} 
& \texttt{compute\_ndbi\_by\_multispectral} & Computes built-up and urban NDBI statistics from a selected multispectral observation. \\
& \texttt{compute\_ndsi\_by\_multispectral} & Computes snow and ice NDSI statistics from a selected multispectral observation. \\
& \texttt{compute\_ndvi\_by\_multispectral} & Computes vegetation NDVI statistics from a selected multispectral observation. \\
& \texttt{compute\_ndwi\_by\_multispectral} & Computes water NDWI statistics from a selected multispectral observation. \\
& \texttt{theme\_index\_lookup} & Recommends the index expression and bands for a theme such as vegetation, water, or urban. \\
\midrule
\multirow{4}{2.0cm}{Relationship and Measurement} 
& \texttt{get\_bbox\_geospatial\_relationship} & Computes coarse relative position between two points or bounding boxes. \\
& \texttt{get\_mask\_geospatial\_relationship} & Compares two binary masks for overlap, containment, or relative spatial relations. \\
& \texttt{normalize\_bounding\_boxes} & Normalizes box coordinates to a shared image scale before comparison or relation analysis. \\
& \texttt{basic\_calculator} & Computes simple arithmetic over counts, areas, ratios, or other numeric outputs. \\
\bottomrule
\end{tabular}
\end{table*}

\subsection{Operation configuration}
\label{operation-config-appendix}

To enable efficient large-scale trajectory synthesis and comprehensively assess agent capabilities, \pipeline{EO-Gym} structures execution across three operational axes. These configurations allow researchers to isolate planning quality from perception noise, evaluate prompt robustness, and test tool-selection capabilities under varying degrees of constraint.

\subsubsection{Unverified vs Verified Mode}
\label{unverified-vs-verified-mode}

The tool response mode dictates whether the agent receives raw tool outputs or ground-truth-assisted observations. This distinction is crucial for separating an agent's reasoning and planning capabilities from the inherent noise of underlying perception models.

When operating in the \textbf{Verified mode}, the environment replaces non-deterministic perception tools with ground-truth-backed simulator outputs. During trajectory synthesis, we do not naively return the dataset ground truth for optical imagery because open-source annotations occasionally exhibit noise, such as oversized bounding boxes or a single box encompassing multiple distinct objects. Instead, the pipeline generates bounding box candidates using optical detection models, specifically Meta SAM 3 \cite{meta2025sam3} and Grounding DINO \cite{liu2023groundingdino}. We cross-validate these predictions against the original ground truth by rendering the bounding boxes onto the images and prompting \model{GPT-4.1-mini} to visually judge and select the most accurate observation. 

For SAR and cross-modal tasks, comparable zero-shot detection foundation models do not currently exist. Consequently, the verified mode relies directly on the ground truth. However, to ensure the tool behaves realistically, we apply a semantic filter. If a tool queries for a specific target like a vehicle but the ground truth only contains car and ship labels, a language model filters the ground truth to return only the relevant car bounding boxes. 

Importantly, the verified mode operates differently during the evaluation phase compared to trajectory synthesis. Because the optical ground truth within the \dataset{EO-Gym-Data} test set synthesized trajectories undergoes rigorous manual verification during the human review phase, the evaluation phase bypasses the visual cross-validation step. Instead, it directly utilizes the verified ground truth coupled with the semantic target filter. To guarantee deterministic and consistent filtering across all evaluation runs, we deploy a dedicated \model{Qwen3-VL-2B-Instruct} \cite{bai2025qwen3vl} model to execute this target-to-ground-truth alignment. This mode eliminates perception noise, guarantees trustworthy observations, and is primarily used for high-quality trajectory generation and isolating planning evaluation.

Conversely, the \textbf{Unverified mode} executes the raw, underlying tools directly without any ground-truth assistance or quality-control filtering. Under Unverified mode, the agent must consume and interpret uncurated optical tool outputs, which may include false positives, missed detections, or misaligned bounding boxes. This setting evaluates end-to-end robustness under realistic noisy EO workflows. In our current implementation, this mode applies only to optical object-related tools, since we do not yet have reliable object-detection backends for SAR imagery or cross-modal object analysis. SAR object-detection tools and cross-modal object-analysis tools are therefore always executed in \textbf{Verified mode}.

\subsubsection{Simple vs Detailed Prompt}
\label{simple-vs-detailed-prompt}

The prompt configuration controls the level of operational guidance provided to the agent during its interaction with the environment. 

The \textbf{Simple prompt} establishes the baseline behavioral expectations. It mandates the execution of at least one task-relevant evidence-gathering tool prior to formulating a final answer. The prompt instructs the agent to treat companion-image retrieval as a setup step rather than final evidence. It also encourages the agent to retry alternative tools if initial results are empty or weak. Furthermore, it explicitly prevents the agent from prematurely terminating its trajectory based on raw detections when addressing complex reasoning tasks such as counting, ratio computation, or change detection. The exact system prompt is provided in Listing \ref{lst:simple_prompt}.

\begin{tcblisting}{
    listing only,              % Tells it to format as code/raw text
    breakable,                 % Allows the box to split across pages
    colback=gray!5!white,      % Light gray background
    colframe=gray!60!black,    % Darker gray border
    title=Listing 1: The Simple Prompt configuration.,
    fonttitle=\bfseries\small, % Bold title
    boxrule=0.5pt,             % Thin border
    arc=4pt,                   % Rounded corners
    left=6pt, right=6pt, top=6pt, bottom=6pt, % Padding
    listing options={
        basicstyle=\small\ttfamily,
        breaklines=true,       % Wraps long lines automatically
        columns=fullflexible   % Prevents weird letter spacing
    },
    label=lst:simple_prompt
}
You are an expert remote-sensing analyst. You must use the given tool to help you see the fine details of the image directly.

Use tools before answering, and ground the final answer in the available observations. 

Stop and answer policy:
- Never give the final answer before at least one task-relevant evidence tool has been used.
- Companion-image lookup is only a setup step, not evidence. A failed lookup is not enough reason to stop.
- If one tool returns an empty result or weak evidence, try another relevant tool before abstaining.
- For counting, ratio, comparison, or change questions, do not stop after raw detections alone when another available tool or reasoning step is needed to convert them into the requested answer.
- Once you already have enough grounded evidence to answer the exact question, stop and answer directly. Do not keep calling extra tools just to reconfirm.
\end{tcblisting}

The \textbf{Detailed prompt} augments the baseline instructions with rigorous, task-specific operational policies. For counting and arithmetic tasks, it requires the agent to actively filter duplicate, partial, or weak detections before aggregation. For ratio and calculation tasks, the agent must explicitly identify the exact numerator and denominator prior to invoking the calculator tool. The prompt also enforces bounding box coordinate verification before deriving areas or spatial splits. To improve robustness, it prevents premature abstention after a single failed detection and introduces cross-modal fallback strategies. If a task requires both SAR and optical imagery, the agent is guided to retrieve the missing companion or continue with the best available modality if the retrieval fails. Finally, it prompts the agent to explicitly verify that it possesses direct tool evidence for the target object or region before generating the final answer. The exact system prompt is provided in Listing \ref{lst:detailed_prompt}.

\begin{tcblisting}{
    listing only,              % Tells it to format as code/raw text
    breakable,                 % Allows the box to split across pages
    colback=gray!5!white,      % Light gray background
    colframe=gray!60!black,    % Darker gray border
    title=Listing 2: The Detailed Prompt configuration.,
    fonttitle=\bfseries\small, % Bold title
    boxrule=0.5pt,             % Thin border
    arc=4pt,                   % Rounded corners
    left=6pt, right=6pt, top=6pt, bottom=6pt, % Padding
    listing options={
        basicstyle=\small\ttfamily,
        breaklines=true,       % Wraps long lines automatically
        columns=fullflexible   % Prevents weird letter spacing
    },
    label=lst:detailed_prompt
}
You are an expert remote-sensing analyst. You must use the given tool to help you see the fine details of the image directly.

Use tools before answering, and ground the final answer in the available observations. Please brief the chain of thought before give the final answer.

Stop and answer policy:
- Never give the final answer before at least one task-relevant evidence tool has been used.
- Companion-image lookup is only a setup step, not evidence. A failed lookup is not enough reason to stop.
- If your current result only says that the paired modality is missing, you must continue with the best tool that works on the available modality.
- If one tool returns an empty result or weak evidence, try another relevant tool before abstaining.
- For counting, ratio, comparison, or change questions, do not stop after raw detections alone when another available tool or reasoning step is needed to convert them into the requested answer.
- Once you already have enough grounded evidence to answer the exact question, stop and answer directly. Do not keep calling extra tools just to reconfirm.

Counting and arithmetic policy:
- For counting questions, the final answer is the number of distinct, identifiable target objects, not automatically the raw number of returned boxes or masks.
- Raw detections are candidate evidence. First interpret them, then answer.
- If detections overlap heavily, look duplicated, are partial at the image boundary, or have weak confidence, do not count all of them blindly.
- If the question says "visible", "identifiable", "fully identifiable", or "distinct", filter out duplicate, partial, and low-confidence detections before answering.
- Do not answer 0 only because one detector call returned an empty list. First try one more relevant evidence tool or a better-targeted prompt if available.
- If the requested image date/time is available, answer from that image. Do not switch to an earlier or fallback image unless the exact requested image is unavailable and you say so explicitly.
- For ratio or arithmetic questions, compute the requested value from the exact categories asked in the question, not from unrelated totals, box counts, or mask areas.
- Before using a calculator, identify exactly which observed quantities are the numerator and denominator and check that they come from the correct target categories.
- If you derive areas or large/small splits from bounding boxes, verify the bbox format before doing arithmetic. Do not assume the coordinate convention incorrectly.
- Before the final answer on counting or arithmetic tasks, verify that the number you will output is the filtered answer the question asks for, not just an intermediate tool statistic.
- Prefer a short evidence summary in your reasoning: observed candidates to filtered valid objects to requested count/ratio/final answer.

Cross-modal fallback policy:
- If a question mentions both SAR and optical imagery, first try to retrieve the missing companion image if only one modality is provided.
- If companion retrieval fails, do not stop just because the paired modality is missing.
- Use the available modality first:
  - If only an optical image is available, use optical-only tools.
  - If only a SAR image is available, use SAR-only tools.
- Use joint optical+SAR tools only when both image paths are available and non-empty.
- Do not answer "cannot determine" only because the paired modality is missing.
- Abstain only when the target cannot be supported by the available modality and its tools.
- If you answer from a single available modality after companion lookup fails, say briefly that the answer is based on the available image evidence.

Practical checklist before the final answer:
1. Did I already call a tool that directly observes the target object, region, change, or scene property?
2. If not, call the best matching available-modality tool now.
3. Is my only reason for stopping that the paired image is missing or the first tool was unhelpful?
4. If yes, do not stop yet. Try the next most relevant available tool.
5. After the last observation, what exact missing uncertainty remains? If one remains, do not answer yet.
6. For counting or ratio questions, did I convert raw detections into the filtered count or calculation the question actually asks for?
7. If I am about to answer from a raw box/mask count, have I checked for duplicates, partial objects, low-confidence detections, and the requested date/modality?
8. If I am about to answer 0 or "cannot determine", do I have enough evidence that the target is truly absent or unsupported rather than merely missed by one detector call?
9. Please brief the chain of thought before give the final answer.
\end{tcblisting}

\subsubsection{Skill vs All Tools}
\label{skill-vs-all-tools}

The tool schema configuration determines the size and relevance of the action space exposed to the agent during execution.

The \textbf{Skill tools} setting narrows the action space by exposing only a dataset-mapped subset of tools relevant to the current task. Rather than presenting the entire catalog, the environment filters the schema based on the underlying data source. This configuration provides a strong prior for tool selection and streamlines the reasoning process. Table \ref{tab:skill-tool-mapping} details the specific tool subsets mapped to each dataset. For instance, multispectral temporal tasks receive a tailored set of tools focused on spectral indices, temporal retrieval, and thematic masking, whereas cross-modal tasks receive a compact set of tools dedicated to optical and SAR switching and paired detection.

The \textbf{All tools} setting exposes the complete registry of $35$ tools to the agent regardless of the underlying dataset or task. In this mode, the full registered tool catalog is exposed to every data source. This configuration is essential for comprehensive all-tool comparison runs, although it intentionally removes the dataset-specific tool prior. It rigorously tests the agent's ability to navigate a massive action space, ignore irrelevant functions, and autonomously select the appropriate data-gathering and analysis tools without relying on predefined constraints.

\begin{table*}[htbp]
\centering
\small
\caption{Dataset-mapped tool subsets under the Skill tools configuration. This mapping restricts the action space to the most relevant functions for each data source.}
\label{tab:skill-tool-mapping}
\begin{tabular}{p{3.0cm}p{1cm}p{9.0cm}}
\toprule
Source Dataset & \# Tool & Main Tool Coverage \\
\midrule
Multispectral (Landsat and Sentinel-2) & 15 & Multispectral listing, previous and next retrieval, multispectral crop, NDVI, NDWI, NDBI, NDSI, water, vegetation, and urban masks, bounding box and mask relationships, calculator, theme lookup. \\
fMoW & 12 & Optical frame list, previous and next RGB retrieval, optical and SAR crop, optical bounding box and mask tools, road and building masks, bounding box and mask relationships, calculator, bounding box normalization. \\
FAIR1M & 11 & Crop, zoom-out, four pan directions, calculator, optical bounding box, bounding box and mask relationships, object mask. \\
DIOR, DOTA, xView & 10 each & Optical crop, optical bounding box and mask tools, road and building masks, bounding box and mask relationships, calculator, scene analysis, object description. \\
xBD & 9 & Optical crop, optical bounding box, object and building masks, bounding box and mask relationships, calculator, scene analysis, object description. \\
M4-SAR & 7 & Optical and SAR bounding box, optical and SAR switching, crop, bounding box and mask relationships, calculator. \\
SARDet-100K & 5 & Crop, SAR bounding box, bounding box relationship, calculator, mask relationship. \\
\bottomrule
\end{tabular}
\end{table*}

\label{sec:operation-confg}

\section{Earth Observation Trajectory Dataset: \dataset{EO-Gym-Data}}

\subsection{\dataset{EO-Gym-Data} Overview}
\label{data-overview-appendix}

The \dataset{EO-Gym-Data} benchmark transforms static EO supervision into interactive, multi-step data-gathering trajectories. To achieve this, we map the original annotations from eight source datasets into a unified taxonomy comprising six primary EO tasks and 18 distinct question types. Furthermore, the environment indexes a massive local data lake that agents can dynamically explore. 

\subsubsection{Indexed Imagery Inventory}

\pipeline{EO-Gym} indexes a comprehensive repository of multimodal files. We categorize the inventory scale into direct and expanded scopes. The direct scope includes files explicitly referenced by the trajectory starting states. The expanded scope encompasses all files reachable through the backend relations of the environment. These include adjacent spatial tiles for panning, historical frames for temporal fetching, and aligned companion pairs for cross-modal switching.

\begin{table*}[htbp]
\centering
\footnotesize
\caption{Imagery inventory counts. The expanded scope represents the full executable data lake available for interactive evidence acquisition.}
\label{tab:imagery-inventory}
\begin{tabular}{ll rrrr}
\toprule
\textbf{Split} & \textbf{Scope} & \textbf{MS Scenes} & \textbf{MS Band Files} & \textbf{RGB Images} & \textbf{SAR Images} \\
\midrule
Train & Direct & 1,346 & 18,160 & 5,632 & 1,809 \\
Train & Expanded & 40,877 & 510,814 & 17,342 & 1,809 \\
Test & Direct & 328 & 4,454 & 823 & 329 \\
Test & Expanded & 10,131 & 127,422 & 3,943 & 384 \\
\midrule
Overall & Direct & 1,674 & 22,614 & 6,455 & 2,138 \\
Overall & Expanded & 51,008 & 638,236 & 21,194 & 2,193 \\
\bottomrule
\end{tabular}
\end{table*}

Table \ref{tab:imagery-inventory} summarizes the imagery inventory available to the agent. The \textit{Direct} inventory counts the exact files named by the trajectory starting states. However, to enable dynamic evidence gathering, the environment expands this into a large indexed backend data lake containing over $660$k files. This \textit{Expanded} scope includes full temporal sequences for fMoW, all regional multispectral captures for FAIR1M, aligned optical and SAR companions for M4-SAR, and high-resolution base images for spatial navigation. 

\subsubsection{Ground Truth Label Distribution.} The ground-truth annotations originate from a robust collection of source datasets, including M4-SAR, SARDet\_100K, DOTA, FAIR1M, xView, DIOR, xBD, and fMoW. These sources provide a rich taxonomy of remote sensing categories. The most prominent specific categories include Road and Surface Transport (45.2\%), Maritime and Offshore (21.4\%), Aircraft and Runways (8.2\%), Buildings and Settlements (4.4\%), and Sports and Leisure (4.0\%), alongside a broad Other category (13.5\%) that captures diverse background contexts. The dataset also incorporates critical disaster assessment and temporal labels, such as Damage and Hazards spanning minor, major, and severe destruction, as well as Change Detection. This diverse label distribution ensures that agents are evaluated on a comprehensive suite of real-world analytical tasks.

\subsubsection{Trajectory and Reasoning Trace Statistics.} Moving beyond static input question-answering, \dataset{EO-Gym-Data} provides extensive multi-step reasoning supervision. The final dataset comprises 9,078 trajectories divided into 7,642 training and 1,436 testing records. These trajectories contain 34,604 reasoning steps and 25,088 function calls. On average, agents execute a median of 3.0 function calls per record, with complex queries requiring up to 46 sequential tool invocations. This depth ensures that models are trained and evaluated on sustained data-gathering workflows. Although the longest synthesized trajectories contain up to 46 tool invocations, the held-out evaluation questions are answerable within the 15-call inference budget; longer traces primarily arise during synthesis and repair loops, where redundant or exploratory calls are retained for training supervision.

\subsubsection{Global Geographical Coverage.} Table \ref{tab:global-coverage} details the geographical distribution of the trajectories across the training and testing splits. The dataset spans all major continents and offshore regions. North America and Asia represent the largest shares, which naturally aligns with the geographical focus of the underlying public source datasets. Importantly, the absolute gap between the training and testing splits remains small across all regions at under 5 percentage points. This balance ensures a consistent and fair evaluation distribution.

\begin{table*}[htbp]
\centering
\footnotesize
\caption{Global geographical coverage of the \dataset{EO-Gym-Data} trajectories, which constitute the only geolocated subset. The distribution maintains consistent regional shares between the training and testing splits.}
\label{tab:global-coverage}
\begin{tabular}{l rrc rrc}
\toprule
\textbf{Region} & \textbf{Train} & \textbf{Train Share} & & \textbf{Test} & \textbf{Test Share} \\
\midrule
North America & 1,755 & 45.7\% & & 422 & 50.4\%  \\
Asia & 996 & 25.9\% & & 205 & 24.5\%  \\
Europe & 591 & 15.4\% & & 127 & 15.2\%  \\
Unmatched/Offshore & 187 & 4.9\% & & 39 & 4.7\%  \\
South America & 147 & 3.8\% & & 22 & 2.6\%  \\
Africa & 113 & 2.9\% & & 10 & 1.2\%  \\
Oceania & 55 & 1.4\% & & 12 & 1.4\%  \\
\bottomrule
\end{tabular}
\end{table*}

\subsubsection{Temporal Coverage.} The dataset features robust temporal dimensions. The temporal reasoning tasks span diverse timeframes, with median observation gaps of 255.5 days for fMoW, 167.0 days for xBD, and 55.0 days for Multispectral sequences. Furthermore, the multispectral subset includes 1,674 curated scenes, predominantly utilizing Sentinel-2 (1,537 scenes) alongside Landsat (137 scenes). To support complex spectral reasoning, the environment integrates raw multispectral data from these multiple satellite platforms. Table \ref{tab:ms-platforms} provides specific details on these multispectral platforms, including Landsat 8, Landsat 9, Sentinel-2A, and Sentinel-2B. The table outlines the exact band files exposed to the agent and their corresponding physical roles. This requires the model to actively select the correct spectral bands to compute indices for water, vegetation, or urban analysis.

\begin{table*}[htbp]
\centering
\footnotesize
\caption{Multispectral platform and sensor details. The environment exposes raw GeoTIFF band files, requiring the agent to understand specific band roles for spectral reasoning.}
\label{tab:ms-platforms}
\resizebox{\textwidth}{!}{%
\begin{tabular}{ll rrr p{4.5cm} p{6cm}}
\toprule
\textbf{Platform} & \textbf{Sensor Family} & \textbf{Scenes} & \textbf{Min Files} & \textbf{Max Files} & \textbf{Observed GeoTIFF Files} & \textbf{Band Roles} \\
\midrule
Landsat 8 & Landsat & 17 & 8 & 8 & SR\_B1.tif, SR\_B2.tif, SR\_B3.tif, SR\_B4.tif, SR\_B5.tif, SR\_B6.tif, SR\_B7.tif, ST\_B10.tif & SR\_B1 coastal aerosol, SR\_B2 blue, SR\_B3 green, SR\_B4 red, SR\_B5 near infrared, SR\_B6 SWIR 1, SR\_B7 SWIR 2, ST\_B10 thermal \\
\addlinespace
Landsat 9 & Landsat & 120 & 8 & 8 & SR\_B1.tif, SR\_B2.tif, SR\_B3.tif, SR\_B4.tif, SR\_B5.tif, SR\_B6.tif, SR\_B7.tif, ST\_B10.tif & SR\_B1 coastal aerosol, SR\_B2 blue, SR\_B3 green, SR\_B4 red, SR\_B5 near infrared, SR\_B6 SWIR 1, SR\_B7 SWIR 2, ST\_B10 thermal \\
\addlinespace
Sentinel-2A & Sentinel-2 & 1,399 & 14 & 14 & B1.tif, B11.tif, B12.tif, B2.tif, B3.tif, B4.tif, B5.tif, B6.tif, B7.tif, B8.tif, B8A.tif, B9.tif, MSK\_CLDPRB.tif, SCL.tif & B1 coastal aerosol, B2 blue, B3 green, B4 red, B5/B6/B7 red edge, B8 near infrared, B8A narrow near infrared, B9 water vapor, B11/B12 SWIR, SCL scene classification, MSK\_CLDPRB cloud probability \\
\addlinespace
Sentinel-2B & Sentinel-2 & 138 & 14 & 14 & B1.tif, B11.tif, B12.tif, B2.tif, B3.tif, B4.tif, B5.tif, B6.tif, B7.tif, B8.tif, B8A.tif, B9.tif, MSK\_CLDPRB.tif, SCL.tif & B1 coastal aerosol, B2 blue, B3 green, B4 red, B5/B6/B7 red edge, B8 near infrared, B8A narrow near infrared, B9 water vapor, B11/B12 SWIR, SCL scene classification, MSK\_CLDPRB cloud probability \\
\bottomrule
\end{tabular}%
}
\end{table*}

\subsubsection{EO Tasks and Question Type Mapping}

The benchmark focuses on six core EO tasks: \textbf{Disaster Impact}, \textbf{Temporal Reasoning}, \textbf{Spatial Navigation}, \textbf{Visual Understanding}, \textbf{Object Counting}, and \textbf{Geospatial Reasoning}. Each task contains specific question types that dictate the required reasoning and tool execution paths. Table \ref{tab:source_to_question_mapping} summarizes how the supervision from each source dataset is utilized to construct these question types. 

For example, the \dataset{xBD} dataset provides pre-event and post-event building polygons alongside damage severity labels. We convert these into Disaster Impact questions such as damage presence, damage count, majority damage, and damage severity. Similarly, the \dataset{fMoW} dataset provides ordered image frames with timestamps, which we map to optical temporal analysis, counting, and overlap analysis. For spatial navigation, we utilize cropped high-resolution windows from FAIR1M to generate moving reasoning questions, forcing the agent to pan and zoom to recover missing context.

\begin{table*}[htbp]
\centering
\small
\caption{Mapping of source datasets to \pipeline{EO-Gym} tasks and question types. The environment converts static annotations into interactive reasoning targets.}
\label{tab:source_to_question_mapping}
\begin{tabular}{p{2.5cm}p{5.5cm}p{5.0cm}}
\toprule
Source Dataset & Source Supervision Utilized & EO-Gym Question Types \\
\midrule
xBD & Before and after building polygons, post-event damage subtypes, and balanced change labels. & \textbf{Disaster Impact:} damage presence, damage count, damage majority, damage severity \\
\midrule
fMoW & Ordered image frames with timestamps and per-frame object bounding boxes. & \textbf{Temporal Reasoning:} optical temporal analysis \newline \textbf{Object Counting:} counting \newline \textbf{Geospatial Reasoning:} area dominance, overlap analysis \\
\midrule
M4-SAR & Co-registered optical and SAR images with object boxes and modality roles. & \textbf{Visual Understanding:} cross-modal analysis \\
\midrule
FAIR1M & Cropped large-image windows with visible object counts and partial-edge object flags. & \textbf{Spatial Navigation:} moving reasoning \\
\midrule
DIOR & Single optical image object categories and bounding boxes. & \textbf{Spatial Navigation:} cut reasoning \newline \textbf{Object Counting:} counting \newline \textbf{Geospatial Reasoning:} orientation, spatial distribution, spatial relationship, overlap analysis \newline \textbf{Visual Understanding:} scene understanding, object attribute \\
\midrule
DOTA & Single optical image object categories and bounding boxes. & \textbf{Spatial Navigation:} cut reasoning \newline \textbf{Object Counting:} counting, arithmetic reasoning \newline \textbf{Geospatial Reasoning:} area dominance, spatial relationship, overlap analysis \newline \textbf{Visual Understanding:} scene understanding \\
\midrule
xView & Single optical image object categories and bounding boxes. & \textbf{Spatial Navigation:} cut reasoning \newline \textbf{Object Counting:} counting, arithmetic reasoning \newline \textbf{Geospatial Reasoning:} spatial relationship, overlap analysis \newline \textbf{Visual Understanding:} scene understanding, object attribute \\
\midrule
SARDet-100K & SAR image object labels and bounding boxes. & \textbf{Spatial Navigation:} cut reasoning \newline \textbf{Object Counting:} arithmetic reasoning \newline \textbf{Geospatial Reasoning:} spatial relationship, overlap analysis, spatial distribution \\
\midrule
GEE Multispectral & Object names paired with multispectral folders, capture dates, sensors, and ground sample distances. & \textbf{Temporal Reasoning:} multispectral temporal analysis \\
\bottomrule
\end{tabular}
\end{table*}

\subsubsection{Train and Test Split}
\label{train-and-test-split-verification}

To ensure robust evaluation, we carefully construct the training and testing splits to prevent data leakage. Strict geographical image isolation is not always feasible because several source datasets, such as DIOR, DOTA, and xView, lack the detailed geolocation metadata required for spatial partitioning. To overcome this limitation, we define our splits based on the complete trajectory context. Each trajectory represents a unique combination of a source image, a specific question type, and a distinct tool execution space. By strictly partitioning these combinations, we successfully guarantee that the training and testing sets remain completely disjoint in their interactive reasoning requirements.

To the datasets possessing precise coordinate metadata, for example the temporal tasks across \dataset{FAIR1M}, \dataset{fMoW}, and \dataset{xBD}, there are exactly zero shared temporal source identifiers between the training and testing splits. For spatial isolation, we performed a leakage-oriented geolocation check by measuring the Haversine distance over WGS84 coordinates between each unique test source and its nearest training counterpart. This analysis focuses on \dataset{fMoW} and \dataset{FAIR1M}, derived records because they provide detailed source geolocation. 

Table \ref{tab:geo-distance-audit} details the results of this nearest-train distance audit. Across the combined 3,488 training and 793 testing unique sources, the median geographical distance is 0.29 kilometers. The \dataset{fMoW} dataset maintains strict spatial separation. It features zero shared source identifiers, and a median distance of 10.32 kilometers, with only 5.1\% of test sources falling within 1 kilometer of a training source. 

In contrast, the \dataset{FAIR1M} family exhibits closer proximity. The GEE Multispectral subset contains zero shared source identifiers but has a median distance of 0.24 kilometers, with 90.2\% of test sources located within 1 kilometer of a training source. The FAIR1M spatial navigation subset contains 35 shared source identifiers and a median distance of 0.48 kilometers. This overlap is an intentional artifact of prompt-variant stratification. Because different reasoning tasks are derived from the same high-resolution base image, the locations may be proximate, but the interactive reasoning trajectories required to solve the tasks remain entirely distinct.

\begin{table*}[htbp]
\centering
\footnotesize
\caption{Nearest-train geolocation distance audit for datasets with precise coordinate metadata. Distance is measured using the Haversine formula over WGS84 coordinates. Shared source identifiers and identical rounded coordinates serve as strict leakage indicators.}
\label{tab:geo-distance-audit}
\begin{tabular}{l rrrrrrr}
\toprule
\textbf{Analysis Scope} & \textbf{Train Sources} & \textbf{Test Sources} & \textbf{Median (km)} & \textbf{P90 (km)} & \textbf{Test $<$1 km} \\
\midrule
fMoW & 1,554 & 277 & 10.32 & 111.11 & 5.1\% \\
GEE Multispectral & 1,346 & 328 & 0.24 & 0.96 & 90.2\% \\
FAIR1M & 588 & 188 & 0.48 & 52.09 & 65.4\% \\
\bottomrule
\end{tabular}
\end{table*}

Furthermore, this trajectory-level formulation naturally mitigates concerns regarding foundational data leakage. While large vision-language models may have encountered the raw images and static ground-truth annotations from these public datasets during their pre-training phases, our benchmark evaluates a fundamentally different capability. \pipeline{EO-Gym} assesses interactive tool execution, multi-step planning, and dynamic evidence acquisition. Because these sequential agent trajectories and specialized tool-use cases are entirely novel and absent from the original static datasets, prior parametric exposure to the raw imagery does not provide a shortcut for the dynamic tool-use evaluation.

\subsection{Candidate Question Answer Construction}
\label{sec:candidate-qa-sample}

\subsubsection{Sampling from Ground Truth}
\label{sampling-from-ground-truth}

To construct a diverse and representative set of candidate questions, the generation pipeline first performs label-driven stratified sampling over the source datasets. For each source record, the pipeline extracts a primary stratification label derived from the ground-truth object categories or bounding boxes. The records are then grouped by this primary label, shuffled using a fixed random seed, and processed in a balanced order. This stratification ensures that the generated candidate pool is not dominated by a few highly frequent object classes or scene types.

Once sampled, the pipeline normalizes the raw ground-truth annotations into a compact, hidden evidence payload tailored to the specific Earth Observation task:
\begin{itemize}
    \item \textbf{Object Detection (DIOR, DOTA, xView, SARDet-100K, M4-SAR):} The pipeline extracts object labels and geometric coordinates to serve as hidden evidence for counting, orientation, overlap, area dominance, and spatial-relation questions. For M4-SAR, it preserves the paired cross-modal context while selecting either the optical or SAR view for the visible prompt.
    \item \textbf{Disaster Impact (xBD):} The pipeline aggregates pre- and post-disaster building counts and damage subtypes, keeping both before and after images in the final question prefix to generate damage presence, count, majority, and severity candidates.
    \item \textbf{Temporal Reasoning (fMoW and GEE Multispectral):} For fMoW, the pipeline sorts frame metadata by timestamp, selects a middle non-extreme image for the visible context, and uses the full frame sequence metadata as hidden evidence. For GEE Multispectral, it filters records with sufficient multispectral index coverage, injects capture dates, ground sample distances, and provider metadata, and provides a selected multispectral folder hint.
    \item \textbf{Spatial Navigation (FAIR1M):} The pipeline utilizes cropped $256 \times 256$ very-high-resolution windows, explicitly filters for crops containing partial-object flags at the boundaries, and records the crop-to-base image mappings to support moving and zoom-navigation questions.
\end{itemize}

\subsubsection{Question Answer Generation}
\label{question-answer-generation}

The normalized hidden evidence payloads are passed to a LLM to generate MCQs. To ensure broad coverage of the taxonomy, the pipeline cycles through dataset-specific prompt variants. The model consumes the hidden evidence and outputs a structured JSON array of MCQs. A subsequent review prompt automatically rewrites or deletes invalid questions before the candidate is finalized and saved.

\textbf{Example 1: GEE Multispectral Temporal Analysis.} The prompt instructs the model to utilize the hidden metadata (capture dates, available multispectral folders, and object lists) to formulate a question about landscape changes over time.

\begin{quote}
\small\ttfamily
Task:\\
- Use the object name list to pick a plausible dominant landscape theme (water, urban, or vegetation) for the current capture.\\
- Ask how that dominant theme changed from the previous capture to the current capture: (A) expanded, (B) contracted, (C) stable, (D) not present in either.\\
\\
Output JSON array with one object, answer="TBD", question\_type="multispectral temporal analysis".
\end{quote}

This instruction generates a candidate question that defers the final answer resolution to the interactive trajectory phase, where the agent must actively compute the spectral indices:

\begin{quote}
\small\ttfamily[Given Multispectral dataset: EO\_GYM\_DATA/e086008c\footnote{The actual URL is EO\_GYM\_DATA/e086008c/e086008cc01208b134ebdde3427ae7d4ea91750215e0e9b938a782586618a11f.} | Capture date: 2020-07-28 | GSD: 10 m/pixel (Sentinel-2)] How has the urban area changed between the previous and current satellite images?\\
Answer: TBD\\
Question\_type: multispectral temporal analysis
\end{quote}

\textbf{Example 2: FAIR1M Spatial Navigation.} The prompt instructs the model to generate a counting question specifically targeting objects that are partially visible at the edge of a cropped window, forcing the agent to navigate to resolve uncertainty.

\begin{quote}
\small\ttfamily
Task:\\
- Select an object category that appears in the current view; prefer one that has at least one partially visible instance.\\
- Ask a "how many objects are visible in this image" question: the total number of that object visible in the current view.\\
- Do NOT mention that the image is cropped/zoomed or use phrases like "cropped patch".\\
- Use partial-object counts to craft plausible distractors.\\
\\
Output JSON array with one object, question\_type="moving reasoning".
\end{quote}

\begin{figure}[htbp]
\centering
\includegraphics[width=0.4\linewidth]{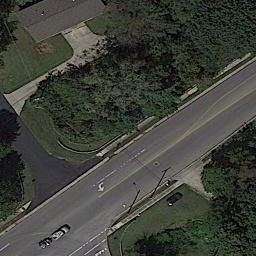}
\caption{An initial cropped observation from the \dataset{FAIR1M} dataset used for spatial navigation generation. The generated question asks the agent to count complete road intersections (the middle bottom has suspicious one), requiring the agent to pan the view to resolve objects partially visible at the image boundaries.}
\label{fig:vhr_zoom_example}
\end{figure}

As illustrated in Figure \ref{fig:vhr_zoom_example}, this instruction yields a candidate question where the agent must actively pan the camera to verify the presence of objects truncated by the initial field of view:

\begin{quote}
\small\ttfamily[Selected image URL: EO\_GYM\_DATA/21e8697d.png\footnote{The actual URL is EO\_GYM\_DATA/21e8697d/21e8697da2007dc3beaa739bb0025cae4d392c85fd1ff3b90bb29e788a05589c.png.}] How many complete road intersections are visible in the image? Please move the view if you need more information to check objects at the edge.\\
Answer: A. 1\\
Question\_type: moving reasoning
\end{quote}

\subsection{Iterative Trajectory Pipeline}
\label{sec:iterative-trajectory-trajectory}

The iterative trajectory pipeline transforms the static question-answer candidates into dynamic tool-use conversations. This process simulates an autonomous agent interacting with the \pipeline{EO-Gym} environment to gather evidence before arriving at a final conclusion.

\subsubsection{Trajectory Generator}
\label{sec:trajectory-generator}

The trajectory generator operates through a multi-step interaction loop. For each candidate record, the generator initializes a conversation with the user's question and the starting image context. The agent (powered by a VLM) processes this prompt and emits a rationale followed by a specific function call. The environment validates the tool arguments, executes the tool (or simulates the observation using the \textbf{Verified mode} mechanisms discussed in Appendix \ref{unverified-vs-verified-mode}), and appends the observation to the conversation history. If a spatial navigation tool is invoked, the environment dynamically updates the visible image crop. This loop continues until the agent emits a final answer or reaches a predefined maximum iteration limit.

To ensure high-quality trajectory synthesis and reduce the search space, we heavily utilize the \textbf{Detailed Prompt} configuration during generation. By injecting strict operational constraints directly into the system prompt, such as enforcing bounding box normalization, dictating exact multispectral discovery sequences, and establishing cross-modal fallback rules—the generator can figure out the correct reasoning trajectory much more easily. Furthermore, the prompt explicitly forbids the agent from hallucinating visual details, forcing it to ground all reasoning strictly in the returned tool observations.

\subsubsection{Trajectory Reviewer}
\label{sec:trajectory-reviewer}

To guarantee the logical coherence of the synthesized trajectories, we implement a rigorous, two-layer review committee that evaluates the trajectory immediately after generation. 

The first layer performs deterministic structural validation. It verifies that the trajectory contains at least one valid tool step, at least one successful non-error observation, utilizes core tools relevant to the dataset, and maintains strictly valid JSON payloads for all function calls.

The second layer employs an LLM-based review committee to audit the logical soundness of the agent's reasoning trace. Importantly, this committee does not compare the agent's answer to the hidden ground truth; instead, it evaluates whether the final conclusion is fully supported by the intermediate tool outputs. The committee explicitly checks for several failure modes:
\begin{itemize}
    \item \textbf{Spatial hallucination:} Drawing directional conclusions from tools that do not provide spatial coordinates.
    \item \textbf{Fabricated inputs:} Passing hardcoded coordinates, empty bounding boxes, or comparing a mask to itself.
    \item \textbf{Evidence contradiction:} Reaching a final conclusion that directly opposes the retrieved tool output.
    \item \textbf{Temporal or multispectral mismatch:} Answering a time-sensitive question without retrieving the requested historical capture, or skipping required neighbor retrieval steps.
\end{itemize}

If the committee rejects a trajectory, it triggers a guidance mechanism. A dedicated feedback generator summarizes the logical failure and proposes a concise, actionable improvement plan. This plan is injected back into the generator as internal feedback, prompting a retry attempt to correct the trajectory.

\subsection{Dataset Enhance and Review}
\label{sec:dataset-enhance-review-appendix}

Following the initial generation and review, the dataset undergoes a comprehensive enhancement stage to clean, standardize, and upgrade the trajectories for optimal model fine-tuning.

\textbf{Rationale Enhancement.} We utilize \model{gpt-oss-120B} rewrite the intermediate assistant rationales preceding each tool call. The enhancer consumes the prior conversation context and the upcoming function call to generate a smooth, concrete justification for the action. This step ensures that the training data demonstrates clear, step-by-step intentionality without leaking future observations or inventing unverified image details.

\textbf{Final-Answer Consistency.} The enhancer also audits the final answer for strict consistency with the accumulated observations and the final thought process. If minor inconsistencies are detected between the reasoning trace and the selected multiple-choice option, the pipeline automatically applies a consistency fix to align the final message with the grounded evidence.

\textbf{MCQ-to-Open-Text Conversion.} To prevent models from relying on process-of-elimination guessing, we deterministically convert the multiple-choice trajectories into an open-text format. The pipeline strips the multiple-choice options from the initial user prompt and rewrites the final answer into a plain-text conclusion. During this conversion, the pipeline also cleans up the conversation history by removing any failed or rejected tool turns, merging the rationales to maintain a seamless, optimal execution path.

A final LLM-driven (\model{gpt-4.1}) rewrite audit is conducted on the open-text trajectories. This step removes residual multiple-choice artifacts (e.g., phrases like "none of the options"), expands overly brief ground-truth answers into complete sentences, and infers missing category details from the tool traces to enrich the user question. Additionally, a specialized check identifies resolution-mismatched questions, such as asking to count small vehicles in 30-meter Landsat imagery, and rewrites the question or trajectory to acknowledge the sensor's physical limitations.

Trajectories that undergo significant LLM rewrites or exhibit borderline audit metadata are automatically flagged and aggregated into a candidate pool for targeted human review, ensuring the highest level of quality control for the final benchmark release.

\subsubsection{Trajectory Quality Control and Review Categories}
\label{sec:trajectory-quality-control}

To guarantee the logical coherence and factual correctness of the synthesized trajectories, we implement a rigorous quality control pipeline. This pipeline evaluates whether the final conclusion is fully supported by the intermediate tool outputs and strictly discards trajectories exhibiting flawed reasoning. The training set undergoes an automated validation process, which filters the initial $7{,}768$ candidates down to $7{,}642$ high-quality trajectories. The test set is subjected to comprehensive human verification, refining $1{,}650$ candidates into a final set of $1{,}436$ trajectories.

During both the automated and human review phases, trajectories are evaluated against a strict set of failure criteria. Table \ref{tab:reviewer_failure_categories} details the specific logical gaps and execution errors that result in trajectory rejection. This strict filtering guarantees that \pipeline{EO-Gym} operationalizes Earth Observation reasoning as a reliable, controlled evidence-acquisition problem.

\begin{table*}[htbp]
\centering
\small
\caption{Reviewer failure categories used during the trajectory quality control phase. Trajectories exhibiting any of these logical gaps are strictly discarded to ensure high-quality reasoning paths.}
\label{tab:reviewer_failure_categories}
\begin{tabular}{p{4.5cm}p{8.5cm}}
\toprule
Reviewer Failure Category & Meaning \\
\midrule
Spatial hallucination & Directional conclusions from tools that do not provide direction. \\
Fabricated inputs & Hardcoded or random coordinates, empty bounding boxes, or comparing a mask to itself. \\
Geometric contradiction & Claimed region or containment result contradicts observed coordinates. \\
Evidence contradiction & Final conclusion directly opposes the tool output. \\
Inference without evidence & Specific attributes are concluded from generic scene descriptions. \\
Temporal mismatch & A timestamp or date question is answered without retrieving the requested capture. \\
Multispectral misuse & Previous, current, or next questions skip required neighbor retrieval after listing captures. \\
No successful observation & All observations are errors or empty. \\
Unclear image input & The input image quality is not clear enough to support reliable object detection during unverified mode execution (identified during test set human review). \\
\bottomrule
\end{tabular}
\end{table*}

\section{Model Fine-Tuning and Evaluation Protocol}

\subsection{Supervised Fine Tuning Setting}
\label{sec:model-fine-tuning}

We develop \model{EO-Gym-4B} by fine-tuning the \model{Qwen3-VL-4B-Instruct} base model using the SWIFT infrastructure. To ensure efficient training, we employ Low-Rank Adaptation (LoRA) targeting all linear modules within the network.  

To strictly enforce active evidence gathering during the training phase, we utilize the Hermes agent template and loss scaling \citep{teknium2024hermes3}, and apply the following fixed system prompt to all training trajectories:

\begin{tcblisting}{
    listing only,              % Tells it to format as code/raw text
    breakable,                 % Allows the box to split across pages
    colback=gray!5!white,      % Light gray background
    colframe=gray!60!black,    % Darker gray border
    title=Listing 3: The Supervised Fine-Tuning (SFT) Prompt configuration.,
    fonttitle=\bfseries\small, % Bold title
    boxrule=0.5pt,             % Thin border
    arc=4pt,                   % Rounded corners
    left=6pt, right=6pt, top=6pt, bottom=6pt, % Padding
    listing options={
        basicstyle=\small\ttfamily,
        breaklines=true,       % Wraps long lines automatically
        columns=fullflexible   % Prevents weird letter spacing
    },
    label=lst:sft_prompt
}
You are an expert remote-sensing analyst. You must use the given tool to help you see the fine details of the image directly. 1. NEVER answer immediately. 2. ALWAYS call a tool first. 3. Only after you are finished with tools, reply on a single line as: Final Answer: <answer>
\end{tcblisting}

We reserve 5\% of the training data for validation splits to monitor convergence. Table~\ref{tab:sft-hyperparameters} details the complete set of hyperparameters used for the supervised fine-tuning process.

\begin{table}[htbp]
\centering
\small
\caption{Hyperparameters for the Supervised Fine-Tuning (SFT) of \model{EO-Gym-4B}.}
\label{tab:sft-hyperparameters}
\begin{tabular}{lc}
\toprule
Hyperparameter & Value \\
\midrule
Training type & LoRA \\
LoRA rank ($r$) & 16 \\
LoRA alpha ($\alpha$) & 32 \\
Target modules & all-linear \\
Number of epochs & 3 \\
Learning rate & 1e-4 \\
Per-device train batch size & 4 \\
Gradient accumulation steps & 16 \\
Max sequence length & 8192 \\
Max image pixels & 1003520 \\
Attention implementation & flash\_attention\_2 \\
Precision & bfloat16 \\
Agent template & hermes \\
Loss scale & hermes \\
\bottomrule
\end{tabular}
\end{table}

\subsection{Inference Infrastructure and Model Versions}
\label{sec:inference_infrastructure}

To ensure reproducibility, we specify the exact model snapshots utilized in our evaluation. The closed-weight models include \model{Gemini-2.5-Pro} (released June 2025), \model{Gemini-2.5-Flash} (released June 2025), \model{GPT-4.1-mini} (gpt-4.1-mini-2025-04-14), \model{GPT-4.1} (gpt-4.1-2025-04-14), and \model{GPT-5.1} (gpt-5.1-2025-11-13).

We tailor the model transport and image handling mechanisms to accommodate the specific requirements of each model family. For the Qwen series, we deploy a local vLLM server with an OpenAI-compatible endpoint. This setup utilizes Hermes tool call parsing and automatic tool choice while directly accessing local file paths for imagery. For the Gemini models, we route requests through the Vertex AI OpenAI-compatible endpoint and encode local images as data URLs. Finally, for the OpenAI models, we utilize the official API, convert local images to HTTP URLs.

\subsection{Evaluation Protocol}
\label{evaluation-protocl-appendix}

\subsubsection{Pass@k Computation}
\label{app:passk}

For closed-weight models, we use dynamic early stopping to reduce API cost. 
If an example is already solved at rollout $j \leq k$, we count it as successful for all larger $k$ and do not generate additional rollouts for that example. 
If no generated rollout among the first $k$ attempts succeeds, the example is counted as unsuccessful for Pass@k. 
We also report 95\% bootstrap confidence intervals by resampling evaluation examples (1,000) and recomputing the dataset-level Pass@k statistic.

\subsubsection{Zero Function Call Trajectories}
\label{sec:evaluation-metrics-appendix}

The core objective of \pipeline{EO-Gym} is to evaluate a model's capacity for interactive evidence acquisition and multi-step tool use. Occasionally, models may output a final answer immediately without invoking any tools, attempting to guess the answer solely from the initial prompt and image. We refer to these instances as \textit{zero function call trajectories}.

Because our benchmark explicitly designs tasks where the initial observation is insufficient to confidently answer the question, a zero-call trajectory indicates a fundamental failure to engage with the environment. More importantly, it is impossible to evaluate a model's data-gathering capacity if it bypasses the tool execution space entirely. Therefore, we define the zero function call trajectories as failure.

Table~\ref{tab:zero-call-rates} details the distribution of zero function call trajectories across all evaluated models under the main benchmark setting (Verified Mode, Skill tools). For the vast majority of models, including our fine-tuned \model{EO-Gym-4B}, the zero-call rate is negligible (well below 4\%). The notable exceptions are the two \model{Qwen3-VL-Thinking} variants, which exhibit zero-call rates exceeding 30\%. This high abstention rate suggests that their internal reasoning mechanisms may prematurely conclude that the task is unsolvable or attempt to bypass external tool use.

\begin{table}[htbp]
\centering
\small
\caption{Distribution of zero function call trajectories across evaluated models corresponding to the main results in Table~\ref{tab:main-simple-prompt}. Models that fail to call any tools are attempting to guess the answer without gathering necessary evidence. To optimize API expenditure for closed weight models, we implement dynamic early stopping during inference. If a model successfully resolves the query in an early rollout, subsequent sampling iterations for that specific instance are bypassed. This dynamic evaluation results in total rollout counts for GPT and Gemini variants being lower than the theoretical maximum of 4,308. For Pass@k under dynamic early stopping, once any rollout among the first $j \le k$ attempts is correct, the example is counted as successful for all larger k and no additional samples are generated; if all generated attempts up to k are incorrect, the example is counted as unsuccessful.}
\label{tab:zero-call-rates}
\begin{tabular}{lccc}
\toprule
Model & \# Trajectories & \# Zero-Call Trajectories & Zero-Call Rate (\%) \\
\midrule
Qwen3-VL-4B-Instruct & 4,308 & 56 & 1.30\% \\
Qwen3-VL-4B-Thinking & 4,308 & 1,550 & 35.98\% \\
Qwen3-VL-8B-Instruct & 4,308 & 3 & 0.07\% \\
Qwen3-VL-8B-Thinking & 4,308 & 1,376 & 31.94\% \\
Qwen3-VL-32B-Instruct & 4,308 & 6 & 0.14\% \\
GPT-4.1-mini & 2,554 & 0 & 0.00\% \\
GPT-4.1 & 2,507 & 1 & 0.04\% \\
GPT-5.1 & 2,598 & 5 & 0.19\% \\
Gemini-2.5-Flash & 2,818 & 44 & 1.56\% \\
Gemini-2.5-Pro & 2,545 & 79 & 3.10\% \\
\midrule
\model{EO-Gym-4B} (Ours) & 4,308 & 1 & 0.02\% \\
\bottomrule
\end{tabular}
\end{table}

\subsubsection{Manual Audit of the Automated Semantic Judge}
\label{sec:manual-audit-judge}

Because our evaluation protocol employs an automated semantic judge (\model{GPT-4.1-mini}), we manually reviewed a balanced sample of $n=600$ model outputs to rigorously assess its reliability. These $600$ evaluation cases were uniformly sampled from the responses generated by all 11 evaluated models (the 10 baseline models and our fine-tuned \model{EO-Gym-4B}) operating under the Verified mode and Skill tools configuration. Each sampled response was compared directly against the corresponding ground-truth answer. To validate our choice of judge model, we also evaluated \model{Gemini-2.5-Flash} and its lightweight variant \model{Gemini-2.5-Flash-Lite} on the identical sample and compared all automated judges against human decisions. 

Each sampled response was compared directly against the corresponding ground-truth answer using the following system instruction: 

\begin{tcblisting}{
    listing only,              % Tells it to format as code/raw text
    breakable,                 % Allows the box to split across pages
    colback=gray!5!white,      % Light gray background
    colframe=gray!60!black,    % Darker gray border
    title=Listing 4: The Semantic Judge Prompt configuration.,
    fonttitle=\bfseries\small, % Bold title
    boxrule=0.5pt,             % Thin border
    arc=4pt,                   % Rounded corners
    left=6pt, right=6pt, top=6pt, bottom=6pt, % Padding
    listing options={
        basicstyle=\small\ttfamily,
        breaklines=true,       % Wraps long lines automatically
        columns=fullflexible   % Prevents weird letter spacing
    },
    label=lst:semantic_judge_prompt
}
You are a strict semantic equivalence judge for QA final answers. Compare REFERENCE and CANDIDATE answers for the same question. Return JSON only: {"is_same_meaning": boolean, "reason": string}. 

Mark false for contradiction or materially different claims. Ignore wording differences if meaning is equivalent.
\end{tcblisting}

Crucially, this evaluation task is highly straightforward and does not require specialized EO background. Human reviewers and automated judges simply compare two text snippets, a concise ground-truth answer and the model's generated response, to determine semantic equivalence. For instance, a response marked as \textit{same} might look like:

\begin{quote}
\small
\textbf{Ground Truth:} The windmill footprint is primarily located in the central part of the scene.\\
\textbf{Model Response:} The windmill footprint is primarily located in the central-right part of the scene within the image. The bounding box for the windmill is roughly between coordinates (444, 408) and (615, 561) on the 800x800 image.
\end{quote}

Conversely, a response marked as \textit{different} might look like:

\begin{quote}
\small
\textbf{Ground Truth:} no-damage\\
\textbf{Model Response:} Based on the analysis of the post-disaster image, the most frequently observed damage category among buildings is "Minor damage". This is evidenced by the presence of floodwaters near the buildings, but with no direct overlap or destruction observed. The relationship between the building masks and the floodwater mask shows no intersection, indicating that while the water is close, it has not inundated the structures themselves.
\end{quote}

As Table \ref{tab:judge-manual-audit-overall} demonstrates, both \model{Gemini-2.5-Flash} and \model{GPT-4.1-mini} achieve exceptional alignment with human reviewers. \model{Gemini-2.5-Flash} yields a marginally higher observed agreement of 0.925 and a Cohen's $\kappa$ of 0.849. \model{GPT-4.1-mini} follows closely with an observed agreement of 0.923 and a $\kappa$ of 0.846. In stark contrast, \model{Gemini-2.5-Flash-Lite} yields a significantly lower observed agreement of 0.775 ($\kappa = 0.537$). The confusion matrix reveals that the \model{Gemini-2.5-Flash-Lite} model is overly conservative. Although it produces only a single false positive, it incorrectly rejects 134 valid paraphrases that human reviewers accepted. Meanwhile, \model{GPT-4.1-mini} and \model{Gemini-2.5-Flash} exhibit much more balanced distributions that align closely with human evaluation standards.

\begin{table}[htbp]
\centering
\small
\caption{Overall pairwise agreement between Human, \model{Gemini-2.5-Flash}, \model{GPT-4.1-mini}, and \model{Gemini-2.5-Flash-Lite} on the manually audited sample of $n=600$ model outputs. $YY$ denotes both raters marking the answer as \textit{same}, $NN$ denotes both marking it as \textit{different}, and $YN$/$NY$ denote disagreement cases.}
\label{tab:judge-manual-audit-overall}
\begin{tabular}{lcccccc}
\toprule
Judge Pair & Obs. Agree. $\uparrow$ & Cohen's $\kappa \uparrow$ & $YY$ & $YN$ & $NY$ & $NN$ \\
\midrule
Human vs. \model{Gemini-2.5-Flash} & 0.925 & 0.849 & 248 & 35 & 10 & 307 \\
Human vs. \model{GPT-4.1-mini} & 0.923 & 0.846 & 263 & 20 & 26 & 291 \\
Human vs. \model{Gemini-2.5-Flash-Lite} & 0.775 & 0.537 & 149 & 134 & 1 & 316 \\
\bottomrule
\end{tabular}
\end{table}

Table \ref{tab:judge-manual-audit-by-bucket} breaks down the agreement across the six EO tasks. Both leading models show near-perfect agreement on discrete tasks like Object Counting, Spatial Navigation, and Temporal Reasoning. However, the agreement drops slightly on Geospatial Reasoning and Disaster Impact. These latter tasks often require interval semantics or directional entailment where a binary prompt struggles to capture human nuance.

\begin{table}[htbp]
\centering
\small
\caption{Agreement between the \model{GPT-4.1-mini} and \model{Gemini-2.5-Flash} semantic judges versus human review across the six primary EO tasks.}
\label{tab:judge-manual-audit-by-bucket}
\begin{tabular}{lccccc}
\toprule
& & \multicolumn{2}{c}{\textbf{\model{GPT-4.1-mini}}} & \multicolumn{2}{c}{\textbf{\model{Gemini-2.5-Flash}}} \\
\cmidrule(lr){3-4} \cmidrule(lr){5-6}
EO Task & $n$ & Agree. $\uparrow$ & $\kappa \uparrow$ & Agree. $\uparrow$ & $\kappa \uparrow$ \\
\midrule
Disaster Impact & 99 & 0.919 & 0.839 & 0.859 & 0.719 \\
Geospatial Reasoning & 102 & 0.853 & 0.700 & 0.863 & 0.712 \\
Object Counting & 101 & 0.970 & 0.940 & 0.980 & 0.960 \\
Spatial Navigation & 100 & 0.940 & 0.876 & 0.980 & 0.959 \\
Temporal Reasoning & 99 & 0.950 & 0.890 & 0.980 & 0.956 \\
Visual Understanding & 99 & 0.909 & 0.816 & 0.889 & 0.780 \\
\bottomrule
\end{tabular}
\end{table}

Ultimately, while \model{Gemini-2.5-Flash} achieves a fractionally higher overall agreement score, \model{GPT-4.1-mini} delivers nearly identical reliability at a significantly lower API cost. This cost efficiency combined with a robust $\kappa$ of $0.846$ strongly justifies using \model{GPT-4.1-mini} for large-scale aggregate reporting. It proves to be the optimal practical choice for evaluating interactive reasoning trajectories, although semantically delicate tasks still benefit from targeted human review.

\section{Experiment}
\label{sec:ablations-appendix}

\subsection{Main Benchmark Result}

\subsubsection{Detailed Tool Execution Metrics}
\label{sec:detailed_tool_metrics}

To comprehensively evaluate the behavioral fidelity of the agents during interactive EO reasoning, we analyze several tool-specific execution metrics. While final answer accuracy measures whether the agent ultimately resolved the query, these tool metrics quantify the efficiency, correctness, and alignment of the agent's evidence-gathering trajectory compared to the optimal ground-truth path.

We report consumed-trajectory tool matching rates to evaluate whether the model used the reference tools during the attempts needed to solve each question under the Simple prompt and Verified mode configuration. For each predicted trajectory, we compute binary tool matching indicators against the reference trajectory:
\begin{itemize}
    \item \textbf{Exact Match:} Checks whether the predicted and reference tool sequences are identical.
    \item \textbf{In-Order Match:} Checks whether the full reference sequence appears as an ordered subsequence.
    \item \textbf{Any-Order Match (Tool-Any):} Checks whether all reference tools appear in the predicted tool multiset.
\end{itemize}

For each question, we average these indicators over the \textit{consumed trajectories}, defined as the attempts evaluated sequentially until the first correct final answer, or all attempts if no correct answer is found. The final reported rate is the mean of the per-question consumed-trajectory rates across the evaluated questions.

Table \ref{tab:detailed_tool_metrics} presents the results across all evaluated models. \model{EO-Gym-4B} achieves the highest performance across all alignment metrics, reaching a 0.593 Exact Match rate and a 0.760 Tool-Any (Any-Order) rate. This demonstrates that trajectory-based fine-tuning not only improves final accuracy but also significantly enhances the operational fidelity of the evidence-gathering process. The agent learns to navigate the environment optimally rather than relying on random exploration.

In contrast, leading generalist models such as \model{GPT-4.1-mini} and \model{Gemini-2.5-Flash} achieve competitive Tool-Any rates (0.697 and 0.694, respectively) but struggle significantly with exact sequence matching (0.365 and 0.374). This indicates that while these models eventually gather the correct evidence, they often take suboptimal, exploratory, or redundant paths to reach the conclusion. Interestingly, \model{GPT-5.1} exhibits surprisingly low tool alignment metrics, including a Tool-Any rate of 0.368. This occurs because the model frequently attempts to bypass the full required tool sequence, relying heavily on its parametric knowledge to guess the answer rather than actively engaging with the environment.

\begin{table*}[htbp]
\centering
\small
\caption{Consumed-trajectory tool execution metrics evaluated under the Simple prompt, Verified mode and Skill tools. \model{EO-Gym-4B} achieves the highest alignment with the optimal evidence-gathering paths.}
\label{tab:detailed_tool_metrics}
\begin{tabular}{lccc}
\toprule
Model & Tool Exact-Match $\uparrow$ & Tool-In-Order $\uparrow$ & Tool-Any $\uparrow$ \\
\midrule
Qwen3-VL-4B-Instruct & 0.424 & 0.565 & 0.572 \\
Qwen3-VL-4B-Thinking & 0.358 & 0.627 & 0.629 \\
Qwen3-VL-8B-Instruct & 0.337 & 0.506 & 0.512 \\
Qwen3-VL-8B-Thinking & 0.382 & 0.617 & 0.619 \\
Qwen3-VL-32B-Instruct & 0.512 & 0.635 & 0.638 \\
GPT-4.1-mini & 0.365 & 0.686 & 0.697 \\
GPT-4.1 & 0.370 & 0.663 & 0.664 \\
GPT-5.1 & 0.355 & 0.367 & 0.368 \\
Gemini-2.5-Flash & 0.374 & 0.683 & 0.694 \\
Gemini-2.5-Pro & 0.382 & 0.605 & 0.617 \\
\midrule
\model{EO-Gym-4B} (Ours) & \textbf{0.593} & \textbf{0.757} & \textbf{0.760} \\
\bottomrule
\end{tabular}
\end{table*}

\subsubsection{Main Benchmark Breakdown by Required Function-Call Count}
\label{sec:function_call_count_breakdown}

This part presents a comprehensive performance breakdown based on the number of reference function calls required by each trajectory. While each trajectory contains a complete interactive trace, including observations, reasoning steps, tool executions, and a final answer, the stratification variable in this analysis is the ground-truth function-call count. We denote this count as $L$. Analyzing performance across different values of $L$ reveals how models handle increasingly complex evidence-gathering workflows.

Table~\ref{tab:prompt-comparison-function-call-count} details the Pass@3 performance of all evaluated models stratified by the required number of reference function calls. We group examples into cases requiring one, two, three, and four or more function calls. The results show that generalist models perform competitively on short workflows that more closely resemble static visual question answering. However, these models exhibit substantial degradation when the required function-call count increases to three or more, highlighting their difficulty with sustained data-gathering reasoning.

Conversely, \model{EO-Gym-4B} maintains robust performance on long-horizon workflows and achieves the highest accuracy for cases requiring three or more function calls under the Simple prompt. Furthermore, while detailed instructions help generalist models mitigate some long-horizon failures, \model{EO-Gym-4B} achieves superior performance on the most interaction-heavy Simple-prompt cases. This suggests that fine-tuning on \dataset{EO-Gym-Data} helps the model internalize the operational rules required for complex EO evidence acquisition.

\begin{table*}[htbp]
\centering
\footnotesize
\caption{Pass@3 performance breakdown by required reference function-call count ($L$). $L$ denotes the number of ground-truth function calls required by a trajectory. While detailed instructions help generalist models on longer workflows, \model{EO-Gym-4B} achieves strong performance using only the Simple prompt, indicating improved internalization of EO tool-use procedures.}
\label{tab:prompt-comparison-function-call-count}
\resizebox{\textwidth}{!}{%
\begin{tabular}{l cccc cccc}
\toprule
& \multicolumn{4}{c}{\textbf{Simple Prompt (Pass@3)}} & \multicolumn{4}{c}{\textbf{Detailed Prompt (Pass@3)}} \\
\cmidrule(lr){2-5} \cmidrule(lr){6-9}
Model & $L=1$ & $L=2$ & $L=3$ & $L\geq4$ & $L=1$ & $L=2$ & $L=3$ & $L\geq4$ \\
\midrule
Qwen3-VL-4B-Instruct & 0.53 & 0.69 & 0.38 & 0.39 & 0.66 & 0.75 & 0.39 & 0.38 \\
Qwen3-VL-4B-Thinking & 0.69 & 0.69 & 0.28 & 0.27 & 0.69 & 0.62 & 0.28 & 0.30 \\
Qwen3-VL-8B-Instruct & 0.58 & 0.67 & 0.35 & 0.37 & 0.70 & 0.76 & 0.44 & 0.45 \\
Qwen3-VL-8B-Thinking & 0.70 & 0.70 & 0.31 & 0.34 & 0.71 & 0.68 & 0.35 & 0.33 \\
Qwen3-VL-32B-Instruct & 0.66 & 0.77 & 0.48 & 0.47 & 0.69 & 0.69 & 0.46 & 0.53 \\
GPT-4.1-mini & 0.74 & \textbf{0.82} & 0.49 & 0.52 & 0.75 & \textbf{0.83} & 0.53 & \textbf{0.70} \\
GPT-4.1 & \textbf{0.77} & 0.75 & 0.49 & 0.52 & 0.75 & 0.75 & 0.47 & 0.66 \\
GPT-5.1 & 0.71 & 0.66 & 0.57 & 0.58 & 0.75 & 0.79 & 0.57 & 0.68 \\
Gemini-2.5-Flash & 0.73 & 0.79 & 0.48 & 0.52 & \textbf{0.76} & 0.79 & 0.53 & 0.66 \\
Gemini-2.5-Pro & 0.76 & 0.79 & 0.55 & 0.58 & 0.68 & 0.76 & 0.57 & 0.62 \\
\midrule
\model{EO-Gym-4B} (Ours) & 0.75 & 0.79 & \textbf{0.71} & \textbf{0.71} & 0.69 & 0.75 & \textbf{0.68} & 0.63 \\
\bottomrule
\end{tabular}%
}
\end{table*}

\subsubsection{Baseline Model Data Leakage Validation by Multispectral Subset}
\label{sec:multispectral_zero_leakage}

A critical concern when evaluating VLMs is the risk of data leakage, where models might have memorized public datasets during their pre-training phase. To definitively waive this risk, we evaluate the models on a strictly isolated \dataset{Multispectral} subset. This subset is newly curated for \pipeline{EO-Gym} and is not derived from existing public VLM benchmark annotations, substantially reducing benchmark-level leakage risk.

Table \ref{tab:multispectral} presents the performance of all models on this subset. The results demonstrate a severe performance degradation across all generalist foundation models. For instance, the strongest baseline model, \model{GPT-5.1}, only achieves a Pass@3 of 0.38 under the Simple prompt and 0.42 under the Detailed prompt. This sharp decline confirms that generalist models struggle to autonomously acquire and synthesize novel multimodal evidence without relying on prior data exposure. 

In contrast, \model{EO-Gym-4B} maintains exceptionally strong performance, achieving a Pass@3 of 0.65 under the Simple prompt and 0.66 under the Detailed prompt. It also executes a highly efficient number of tool calls while maintaining the highest tool selection accuracy. This provides conclusive evidence that \model{EO-Gym-4B}'s superior performance is not an artifact of data leakage. Instead, it proves that our trajectory fine-tuning successfully teaches the model the fundamental operational rules required to actively gather and reason over unseen Earth Observation data.

\begin{table*}[htbp]
\centering
\footnotesize
\caption{Performance on the \dataset{Multispectral} subset. Generalist models suffer significant performance drops on this entirely novel data. \model{EO-Gym-4B} achieves superior performance using only the Simple prompt, proving its capabilities stem from robust interactive reasoning rather than dataset memorization.}
\label{tab:multispectral}
\resizebox{\textwidth}{!}{%
\begin{tabular}{l cccc cccc}
\toprule
& \multicolumn{4}{c}{\textbf{Simple Prompt}} & \multicolumn{4}{c}{\textbf{Detailed Prompt}} \\
\cmidrule(lr){2-5} \cmidrule(lr){6-9}
Model & P@1 $\uparrow$ & P@3 $\uparrow$ & TC ratio $\downarrow$ & Tool-any $\uparrow$ & P@1 $\uparrow$ & P@3 $\uparrow$ & TC ratio $\downarrow$ & Tool-any $\uparrow$ \\
\midrule
Qwen3-VL-4B-Instruct & 0.16 & 0.21 & 5.05 & 0.31 & 0.16 & 0.21 & 4.18 & 0.23 \\
Qwen3-VL-4B-Thinking & 0.04 & 0.06 & 2.53 & 0.17 & 0.06 & 0.09 & 1.56 & 0.00 \\
Qwen3-VL-8B-Instruct & 0.20 & 0.24 & 3.91 & 0.33 & 0.23 & 0.27 & 3.79 & 0.18 \\
Qwen3-VL-8B-Thinking & 0.05 & 0.07 & 3.25 & 0.42 & 0.07 & 0.10 & 2.21 & 0.08 \\
Qwen3-VL-32B-Instruct & 0.25 & 0.30 & 5.19 & 0.31 & 0.24 & 0.29 & 4.21 & 0.38 \\
GPT-4.1-mini & 0.20 & 0.25 & 5.99 & 0.32 & 0.21 & 0.25 & 6.08 & 0.30 \\
GPT-4.1 & 0.24 & 0.29 & 5.21 & 0.10 & 0.15 & 0.19 & 5.71 & 0.14 \\
GPT-5.1 & 0.33 & 0.38 & 4.64 & 0.03 & 0.37 & 0.42 & 4.96 & 0.11 \\
Gemini-2.5-Flash & 0.23 & 0.27 & 5.18 & 0.16 & 0.22 & 0.26 & 4.87 & 0.11 \\
Gemini-2.5-Pro & 0.29 & 0.34 & 4.93 & 0.19 & 0.27 & 0.32 & 4.08 & 0.16 \\
\midrule
\model{EO-Gym-4B} (Ours) & \textbf{0.60} & \textbf{0.65} & \textbf{3.08} & \textbf{0.66} & \textbf{0.61} & \textbf{0.66} & \textbf{2.95} & \textbf{0.63} \\
\bottomrule
\end{tabular}%
}
\end{table*}

\subsection{RQ1: End-to-end Evaluation}

To rigorously assess end-to-end robustness, we compare model performance across Verified Skill-tool and Unverified All-tool settings. Verified mode isolates planning capability by providing ground-truth observations and restricting the action space to task-relevant tools. In contrast, Unverified All-tool mode exposes the agent to raw perception noise and the complete, unconstrained catalog of 35 tools.

Table~\ref{tab:verified-vs-unverified} reports the performance shifts for five representative models under the Simple prompt configuration. The results show a clear divergence between final-answer accuracy and tool-use behavior. Pass@1 remains relatively stable across settings, with an average change of only $-1.2$ percentage points. Some larger models, including \model{Qwen3-VL-8B-Instruct} and \model{Qwen3-VL-32B-Instruct}, even show small Pass@1 gains under the Unverified All-tool setting.

However, this stable final-answer accuracy masks a substantial drop in tool-use behavior. Tool-any decreases for every evaluated model when moving from Verified Skill-tool mode to Unverified All-tool mode. \model{EO-Gym-4B} remains the strongest model overall, achieving the best Unverified Pass@1 of $0.61$ and the highest Unverified Tool-any rate of $0.456$. Nevertheless, its Tool-any rate drops by $30.4$ percentage points, from $0.760$ to $0.456$, showing that the full raw execution space makes evidence acquisition more difficult even for the fine-tuned model. These results illustrate why final-answer accuracy alone is insufficient for Earth Observation agent evaluation and highlight the need for \model{EO-Gym}'s executable trajectory-based assessment.

\begin{table*}[htbp]
\centering
\footnotesize
\caption{Comparison of Verified Skill-tool mode and Unverified All-tool mode under the Simple prompt. Final-answer accuracy remains relatively stable, while tool-use behavior, measured by Tool-any, drops substantially when models must operate over all 35 tools with raw outputs.}
\label{tab:verified-vs-unverified}
\resizebox{\textwidth}{!}{%
\begin{tabular}{l ccc ccc}
\toprule
& \multicolumn{3}{c}{\textbf{Pass@1 Accuracy}} & \multicolumn{3}{c}{\textbf{Tool-any Rate}} \\
\cmidrule(lr){2-4} \cmidrule(lr){5-7}
Model & Verified & Unverified & $\Delta$ & Verified & Unverified & $\Delta$ \\
\midrule
Qwen3-VL-4B-Instruct  & 0.42 & 0.39 & $-$0.03 & 0.572 & 0.429 & $-$0.143 \\
Qwen3-VL-8B-Instruct  & 0.45 & 0.47 & $+$0.02 & 0.512 & 0.423 & $-$0.089 \\
Qwen3-VL-32B-Instruct & 0.51 & 0.53 & $+$0.02 & 0.638 & 0.409 & $-$0.229 \\
GPT-4.1-mini          & 0.56 & 0.53 & $-$0.03 & 0.697 & 0.429 & $-$0.268 \\
\midrule
\model{EO-Gym-4B} (Ours) & \textbf{0.65} & \textbf{0.61} & $-$0.04 & \textbf{0.760} & \textbf{0.456} & $-$0.304 \\
\bottomrule
\end{tabular}%
}
\end{table*}

A qualitative analysis of specific reasoning trajectories further illuminates the impact of perception noise during unverified execution. When exposed to raw detector outputs, agents frequently misinterpret missing bounding boxes as the definitive absence of objects. Table \ref{tab:qualitative_unverified_failures} presents representative cases where the unverified mode fails due to perception errors, whereas the verified mode successfully resolves the query. For instance, in disaster assessment tasks, the failure to detect post-event buildings leads the unverified agent to erroneously conclude severe destruction. Conversely, the verified mode provides reliable mask intersection over union metrics, enabling the agent to correctly identify intact structures. Similar perception-driven reasoning failures occur in spatial and temporal tasks, where noisy optical detections or retrieval errors derail the evidence-gathering process.

\begin{table*}[htbp]
\centering
\footnotesize
\caption{Qualitative examples illustrating reasoning failures caused by perception noise in the Unverified mode compared to successful evidence acquisition in the Verified mode.}
\label{tab:qualitative_unverified_failures}
\begin{tabular}{p{2cm} p{5.5cm} p{5cm}}
\toprule
\textbf{Task Context} & \textbf{Unverified Execution Failure} & \textbf{Verified Execution Success} \\
\midrule
Disaster Impact & Counts 35 pre-event buildings versus 6 post-event buildings, erroneously estimating a severe damage ratio of 83 percent. & Accurately counts 61 pre-event versus 58 post-event buildings, correctly placing the damage ratio within the 0 to 25 percent range. \\
\addlinespace
Geospatial Reasoning & Relies on noisy optical bounding box detections, leading the agent to incorrectly classify the dominant area as a recreational facility. & Leverages precise mask results to correctly determine that a residential area covers the largest visible portion of the image. \\
\addlinespace
Temporal Reasoning & Prematurely concludes that the exact requested historical image is unavailable and fails to answer the query. & Successfully retrieves the requested timestamp image and accurately detects the target educational institution. \\
\bottomrule
\end{tabular}
\end{table*}

\subsection{RQ2: Tool \& transfer evaluation}

\subsubsection{Tool Group Splitting Methodology}
\label{sec:tool_group_splitting}

To evaluate this true generalization, we partition the tool space into disjoint groups for our transfer experiments.

For \textbf{Cross-modal Switching (CMS)}, the split is strictly directional and clean. The training set exclusively uses the \texttt{get\_sar\_from\_optical} bridge family to gather evidence, while the test set exclusively relies on \texttt{get\_optical\_from\_sar}. The only shared functions between the splits are basic support tools, specifically \texttt{basic\_calculator}, \texttt{get\_bbox\_geospatial\_relationship}, and \texttt{get\_object\_bbox\_by\_optical\_sar\_image}. Consequently, there is zero exact sequence overlap and zero tool set overlap between the training and testing phases.

For \textbf{Temporal Fetching (TF)}, we enforce a clean modality split. All 764 training trajectories exclusively contain optical temporal tools. Conversely, all 265 test trajectories rely entirely on multispectral temporal tools. The training set contains no multispectral temporal tools, and the test set contains no optical temporal tools. This strict isolation results in zero shared individual tools, zero exact tool sequences, and zero tool sets between the splits.

For \textbf{Spatial Planning (SP)}, the setup evaluates question type transfer rather than a strict tool level holdout. The training and test sets share 9 tools and 7 exact tool sequences. These shared exact sequences cover 109 out of the 114 test trajectories. The primary reused patterns consist of single tool or near single tool trajectories, such as \texttt{get\_object\_bbox\_by\_optical\_image}, \texttt{get\_object\_bbox\_by\_sar\_image}, and the sequential execution of \texttt{get\_optical\_image\_list} followed by \texttt{get\_object\_bbox\_by\_optical\_image}. This demonstrates that while the model successfully transfers knowledge to new spatial reasoning questions, the underlying tool execution paths share significant overlap with the training distribution.

\subsubsection{Challenges in Strict Tool-Level Isolation}
\label{sec:tool_disjoint_challenges}

To evaluate the zero-shot tool transfer capabilities of the models, we initially explored enforcing a strict tool-disjoint split by holding out entire EO tasks or specific question types. However, interactive EO reasoning is highly compositional. Foundational operations, such as object detection, mask generation, and arithmetic calculations, serve as universal building blocks across diverse analytical workflows. 

Table \ref{tab:bucket_leave_one_out} presents the results of a leave-one-out analysis at the primary task level. When holding out any of the six primary EO tasks, 100 percent of the test trajectories still rely on tools that are actively used in the remaining training tasks. Ubiquitous tools like \texttt{get\_object\_bbox\_by\_optical\_image}, \texttt{get\_building\_mask\_by\_optical\_image}, and \texttt{basic\_calculator} bridge virtually every task category. 

\begin{table*}[htbp]
\centering
\footnotesize
\caption{Task-level leave-one-out tool analysis. Holding out an entire primary task fails to isolate the underlying tools, as 100 percent of test records continue to rely on foundational tools shared with the training distribution.}
\label{tab:bucket_leave_one_out}
\begin{tabular}{l c c c}
\toprule
\textbf{Held-out Test Task} & \textbf{Test Records} & \textbf{Shared Tools} & \textbf{Shared Tool Usage} \\
\midrule
Disaster Impact & 44 & 3 & 100.0\% \\
\addlinespace
Temporal Reasoning & 384 & 11 & 100.0\%  \\
\addlinespace
Spatial Navigation & 281 & 7 & 100.0\% \\
\addlinespace
Visual Understanding & 362 & 14 & 100.0\%  \\
\addlinespace
Object Counting & 129 & 8 & 100.0\%  \\
\addlinespace
Geospatial Reasoning & 236 & 13 & 100.0\%  \\
\bottomrule
\end{tabular}
\end{table*}

This ubiquitous tool reuse explains why we do not apply a global strict tool-disjoint split across the entire benchmark. Enforcing strict isolation would necessitate discarding any trajectory containing a shared tool. Because these foundational tools are present in virtually every workflow, this filtering process would decimate the dataset size. It would leave an insufficient number of trajectories for meaningful model fine-tuning or robust evaluation. Consequently, to evaluate tool transferability, we rely on the carefully constructed targeted disjoint subsets detailed in our main experiments rather than a global task-level tool holdout.

\subsection{RQ3: Tool Renaming}
\label{sec:tool_renaming_protocol}

To rigorously assess whether the agent relies on memorized tool identifiers rather than underlying functional semantics, we introduce a systematic tool renaming protocol during inference. Under this protocol, only the initial underscore-delimited token of each tool name is substituted with a semantically equivalent alias. This renaming strategy ensures that the evaluated interactive Earth Observation reasoning reflects a genuine understanding of the evidence-acquisition process across space, time, and modality, rather than superficial pattern matching.

Table \ref{tab:rename_rules} details the exact mapping rules applied to the first token of the tool names. Table \ref{tab:rename_examples} provides representative examples illustrating how the backend tool names are transformed into the novel model-facing tool names presented to the agent.

\begin{table}[htbp]
\centering
\small
\caption{Mapping of original tool name prefixes to their model-facing aliases during the renaming evaluation.}
\label{tab:rename_rules}
\begin{tabular}{ll}
\toprule
\textbf{Original First Token} & \textbf{Model-Facing Alias} \\
\midrule
\texttt{get} & \texttt{access} \\
\texttt{compute} & \texttt{derive} \\
\texttt{analyze} & \texttt{inspect} \\
\texttt{describe} & \texttt{characterize} \\
\texttt{crop} & \texttt{clip} \\
\texttt{move} & \texttt{shift} \\
\texttt{zoom} & \texttt{widen} \\
\texttt{normalize} & \texttt{standardize} \\
\texttt{theme} & \texttt{topic} \\
\texttt{basic} & \texttt{simple} \\
\bottomrule
\end{tabular}
\end{table}

\begin{table}[htbp]
\centering
\small
\caption{Representative examples illustrating the transformation from backend tool names to model-facing tool names under the renaming protocol.}
\label{tab:rename_examples}
\begin{tabular}{ll}
\toprule
\textbf{Original Tool Name} & \textbf{Renaming Tool Name} \\
\midrule
\texttt{get\_object\_bbox\_by\_sar\_image} & \texttt{access\_object\_bbox\_by\_sar\_image} \\
\texttt{compute\_vegetation\_mask} & \texttt{derive\_vegetation\_mask} \\
\texttt{analyze\_optical\_scene} & \texttt{inspect\_optical\_scene} \\
\texttt{crop\_optical\_or\_sar\_image} & \texttt{clip\_optical\_or\_sar\_image} \\
\texttt{basic\_calculator} & \texttt{simple\_calculator} \\
\bottomrule
\end{tabular}
\end{table}


\begin{thebibliography}{99}
\small

\bibitem{joshi2016opticalsar}
Joshi, N., Baumann, M., Ehammer, A., Fensholt, R., Grogan, K., Hostert, P., Jepsen, M.R., Kuemmerle, T., Meyfroidt, P., Mitchard, E.T.A., Reiche, J., Ryan, C.M., and Waske, B. (2016).
A review of the application of optical and radar remote sensing data fusion to land use mapping and monitoring.
{\it Remote Sensing}, {\bf 8}(1):70.
\url{https://doi.org/10.3390/rs8010070}.

\bibitem{claverie2018hls}
Claverie, M., Ju, J., Masek, J.G., Dungan, J.L., Vermote, E.F., Roger, J.-C., Skakun, S.V., and Justice, C. (2018).
The harmonized Landsat and Sentinel-2 surface reflectance data set.
{\it Remote Sensing of Environment}, {\bf 219}:145--161.
\url{https://doi.org/10.1016/j.rse.2018.09.002}.

\bibitem{ju2025hlsv2}
Ju, J., Zhou, Q., Freitag, B., Roy, D.P., Zhang, H.K., Sridhar, M., Mandel, J., Arab, S., Schmidt, G., Crawford, C.J., Gascon, F., Strobl, P.A., Masek, J.G., and Neigh, C.S.R. (2025).
The harmonized Landsat and Sentinel-2 Version 2.0 surface reflectance dataset.
{\it Remote Sensing of Environment}, {\bf 324}:114723.
\url{https://doi.org/10.1016/j.rse.2025.114723}.

\bibitem{wulder2016landsat}
Wulder, M.A., White, J.C., Loveland, T.R., Woodcock, C.E., Belward, A.S., Cohen, W.B., Fosnight, E.A., Shaw, J., Masek, J.G., and Roy, D.P. (2016).
The global Landsat archive: Status, consolidation, and direction.
{\it Remote Sensing of Environment}, {\bf 185}:271--283.
\url{https://doi.org/10.1016/j.rse.2015.11.032}.


\bibitem{liu2024llavaplus}
Liu, S., Cheng, H., Liu, H., Zhang, H., Li, F., Ren, T., Zou, X., Yang, J., Su, H., Zhu, J., Zhang, L., Gao, J., and Li, C. (2024).
LLaVA-Plus: Learning to Use Tools for Creating Multimodal Agents.
In {\it Computer Vision -- ECCV 2024}, pp.\ 126--142.
\url{https://doi.org/10.1007/978-3-031-72970-6_8}.

\bibitem{yao2023react}
Yao, S., Zhao, J., Yu, D., Du, N., Shafran, I., Narasimhan, K. R., and Cao, Y. (2023).
ReAct: Synergizing reasoning and acting in language models.
In {\it The Eleventh International Conference on Learning Representations}.
\url{https://openreview.net/forum?id=WE_vluYUL-X}.

\bibitem{qin2024toolllm}
Qin, Y., Liang, S., Ye, Y., Zhu, K., Yan, L., Lu, Y., Lin, Y., Cong, X., Tang, X., Qian, B., Zhao, S., Hong, L., Tian, R., Xie, R., Zhou, J., Gerstein, M., Li, D., Liu, Z., and Sun, M. (2024).
ToolLLM: Facilitating large language models to master 16000+ real-world APIs.
In {\it The Twelfth International Conference on Learning Representations}.
\url{https://openreview.net/forum?id=dHng2O0Jjr}.

\bibitem{koh2024visualwebarena}
Koh, J.Y., Lo, R., Jang, L., Duvvur, V., Lim, M., Huang, P.-Y., Neubig, G., Zhou, S., Salakhutdinov, R., and Fried, D. (2024).
VisualWebArena: Evaluating Multimodal Agents on Realistic Visually
Grounded Web Tasks.
In {\it Proceedings of the 62nd Annual Meeting of the Association for Computational Linguistics (Volume 1: Long Papers)}, pp.\ 881--905.
\url{https://doi.org/10.18653/v1/2024.acl-long.50}.

\bibitem{xie2024osworld}
Xie, T., Zhang, D., Chen, J., Li, X., Zhao, S., Cao, R., Hua, T.J., Cheng, Z., Shin, D., Lei, F., Liu, Y., Xu, Y., Zhou, S., Savarese, S., Xiong, C., Zhong, V., and Yu, T. (2024).
OSWorld: Benchmarking multimodal agents for open-ended tasks in real computer environments.
In {\it Advances in Neural Information Processing Systems}, {\bf 37}:52040--52094.
\url{https://doi.org/10.52202/079017-1650}.

\bibitem{zhang2024earthgpt}
Zhang, W., Cai, M., Zhang, T., Zhuang, Y., and Mao, X. (2024).
EarthGPT: A Universal Multimodal Large Language Model for Multisensor Image Comprehension in Remote Sensing Domain.
{\it IEEE Transactions on Geoscience and Remote Sensing}, {\bf 62}:1--20.
\url{https://doi.org/10.1109/TGRS.2024.3409624}.

\bibitem{zhan2025skyeyegpt}
Zhan, Y., Xiong, Z., and Yuan, Y. (2025).
SkyEyeGPT: Unifying remote sensing vision-language tasks via instruction tuning with large language model.
{\it ISPRS Journal of Photogrammetry and Remote Sensing}, {\bf 221}:64--77.
\url{https://doi.org/10.1016/j.isprsjprs.2025.01.020}.

\bibitem{li2024vrsbench}
Li, X., Ding, J., and Elhoseiny, M. (2024).
VRSBench: A versatile vision-language benchmark dataset for remote sensing image understanding.
In {\it Advances in Neural Information Processing Systems}, {\bf 37}:3229--3242.
\url{https://doi.org/10.52202/079017-0106}.

\bibitem{lacoste2023geobench}
Lacoste, A., Lehmann, N., Rodriguez, P., Sherwin, E., Kerner, H., L{"u}tjens, B., Irvin, J., Dao, D., Alemohammad, H., Drouin, A., Gunturkun, M., Huang, G., Vazquez, D., Newman, D., Bengio, Y., Ermon, S., and Zhu, X. (2023).
GEO-Bench: Toward foundation models for Earth monitoring.
In {\it Advances in Neural Information Processing Systems 36}.
\url{https://doi.org/10.52202/075280-2223}.

\bibitem{soni2025earthdial}
Soni, S., Dudhane, A., Debary, H., Fiaz, M., Munir, M.A., Danish, M.S., Fraccaro, P., Watson, C.D., Klein, L.J., Khan, F.S., and Khan, S. (2025).
EarthDial: Turning multi-sensory Earth observations to interactive dialogues.
In {\it Proceedings of the IEEE/CVF Conference on Computer Vision and Pattern Recognition}, pp.\ 14303--14313.
\url{https://doi.org/10.1109/CVPR52734.2025.01334}.

\bibitem[Kao et al.(2025)]{kao2025univearth}
Kao, C.-H., Zhao, W., Revankar, S., Speas, S., Bhagat, S., Datta, R., Phoo, C.P., Mall, U., Vondrick, C., Bala, K., and Hariharan, B. (2025).
Towards LLM Agents for Earth Observation.
arXiv preprint arXiv:2504.12110.
\url{https://doi.org/10.48550/arXiv.2504.12110}.

\bibitem{shabbir2025thinkgeo}
Shabbir, A., Munir, M.A., Dudhane, A., Sheikh, M.U., Khan, M.H., Fraccaro, P., Moreno, J.B., Khan, F.S., and Khan, S. (2025).
ThinkGeo: Evaluating Tool-Augmented Agents for Remote Sensing Tasks.
arXiv preprint arXiv:2505.23752.
\url{https://doi.org/10.48550/arXiv.2505.23752}.

\bibitem{feng2026earthagent}
Feng, P., Lv, Z., Ye, J., Wang, X., Huo, X., Yu, J., Xu, W., Zhang, W., Bai, L., He, C., and Li, W. (2026).
Earth-Agent: Unlocking the Full Landscape of Earth Observation with Agents.
In {\it The Fourteenth International Conference on Learning Representations}.

\bibitem{shabbir2026openearthagent}
Shabbir, A., Munir, M.A., Sheikh, M.U., Hussain, S., Khan, M.H., Fraccaro, P., Moreno, J.B., Khan, F.S., and Khan, S. (2026).
OpenEarthAgent: A Unified Framework for Tool-Augmented Geospatial Agents.
\url{https://doi.org/10.48550/arXiv.2602.17665}.

\bibitem{zhao2026openearthagent}
Zhao, S., Liu, F., Zhang, X., Chen, H., Gu, X., Jiang, Z., Ling, F., Fei, B., Zhang, W., Wang, J., Xuan, W., Xiao, P., Yokoya, N., and Bai, L. (2026).
OpenEarth-Agent: From Tool Calling to Tool Creation for Open-Environment Earth Observation.
\url{https://doi.org/10.48550/arXiv.2603.22148}.

\bibitem{brockman2016gym}
Brockman, G., Cheung, V., Pettersson, L., Schneider, J., Schulman, J., Tang, J., and Zaremba, W. (2016).
OpenAI Gym.
\url{https://doi.org/10.48550/arXiv.1606.01540}.

\bibitem{drusch2012sentinel}
Drusch, M., Del Bello, U., Carlier, S., Colin, O., Fernandez, V., Gascon, F., Hoersch, B., Isola, C., Laberinti, P., Martimort, P., Meygret, A., Spoto, F., Sy, O., Marchese, F., and Bargellini, P. (2012).
Sentinel-2: ESA's optical high-resolution mission for GMES operational services.
{\it Remote Sensing of Environment}, {\bf 120}:25--36.
\url{https://doi.org/10.1016/j.rse.2011.11.026}.

\bibitem{torres2012sentinel1}
Torres, R., Snoeij, P., Geudtner, D., Bibby, D., Davidson, M., Attema, E., Potin, P., Rommen, B., Floury, N., Brown, M., Traver, I.N., Deghaye, P., Duesmann, B., Rosich, B., Miranda, N., Bruno, C., L'Abbate, M., Croci, R., Pietropaolo, A., Huchler, M., and Rostan, F. (2012).
GMES Sentinel-1 mission.
{\it Remote Sensing of Environment}, {\bf 120}:9--24.
\url{https://doi.org/10.1016/j.rse.2011.05.028}.

\bibitem{digitalglobe2014worldview3}
DigitalGlobe. (2014).
WorldView-3 Data Sheet.
DigitalGlobe.
\url{https://www.spaceimagingme.com/downloads/sensors/datasheets/DG_WorldView3_DS_2014.pdf}.
Accessed: May 1, 2026.

\bibitem{toutin2002quickbird}
Toutin, T., and Cheng, P. (2002).
QuickBird---A Milestone for High Resolution Mapping.
{\it Earth Observation Magazine}, {\bf 11}(4):14--18.

\bibitem{madden2009geoeye1}
Madden, M. (2009).
GeoEye-1, the World's highest resolution commercial satellite.
In {\it Conference on Lasers and Electro-Optics/International Quantum Electronics Conference}, OSA Technical Digest (CD), paper PWB4.
\url{https://doi.org/10.1364/CLEO.2009.PWB4}.

\bibitem{huang2018gf2}
Huang, W., Sun, S., Jiang, H., Gao, C., and Zong, X. (2018).
GF-2 Satellite 1m/4m Camera Design and In-Orbit Commissioning.
{\it Chinese Journal of Electronics}, {\bf 27}(6):1316--1321.
\url{https://doi.org/10.1049/cje.2018.09.018}.

\bibitem{wmo2026jilin1}
World Meteorological Organization. (2026).
Satellite: Jilin-1.
OSCAR: Observing Systems Capability Analysis and Review Tool.
\url{https://space.oscar.wmo.int/satellites/view/jilin_1}.
Accessed: May 1, 2026.

\bibitem{cyclomedia2026aerial}
Cyclomedia. (2026).
Aerial data.
\url{https://www.cyclomedia.com/en/producten/data-visualisatie/aerial-data/}.
Accessed: May 1, 2026.

\bibitem{mialon2024gaia}
Mialon, G., Fourrier, C., Wolf, T., LeCun, Y., and Scialom, T. (2024).
GAIA: A benchmark for general AI assistants.
In {\it The Twelfth International Conference on Learning Representations}.
\url{https://openreview.net/forum?id=fibxvahvs3}.

\bibitem{wang2024gta}
Wang, J., Ma, Z., Li, Y., Zhang, S., Chen, C., Chen, K., and Le, X. (2024).
GTA: A benchmark for general tool agents.
In {\it Advances in Neural Information Processing Systems}, {\bf 37}:75749--75790.
\url{https://doi.org/10.52202/079017-2412}.

\bibitem{nathani2025mlgym}
Nathani, D., Madaan, L., Roberts, N., Bashlykov, N., Menon, A., Moens, V., Budhiraja, A., Magka, D., Vorotilov, V., Chaurasia, G., Hupkes, D., Cabral, R.S., Shavrina, T., Foerster, J., Bachrach, Y., Wang, W.Y., and Raileanu, R. (2025).
MLGym: A new framework and benchmark for advancing AI research agents.
\url{https://doi.org/10.48550/arXiv.2502.14499}.

\bibitem{jain2025r2egym}
Jain, N., Singh, J., Shetty, M., Zheng, L., Sen, K., and Stoica, I. (2025).
R2E-Gym: Procedural environments and hybrid verifiers for scaling open-weights SWE agents.
\url{https://doi.org/10.48550/arXiv.2504.07164}.

\bibitem{zhou2024webarena}
Zhou, S., Xu, F.F., Zhu, H., Zhou, X., Lo, R., Sridhar, A., Cheng, X., Ou, T., Bisk, Y., Fried, D., Alon, U., and Neubig, G. (2024).
WebArena: A realistic web environment for building autonomous agents.
In {\it The Twelfth International Conference on Learning Representations}.
\url{https://openreview.net/forum?id=oKn9c6ytLx}.

\bibitem{li2025formfactory}
Li, B., Wang, Y., Fei, H., Li, J., Ji, W., Lee, M.-L., and Hsu, W. (2025).
FormFactory: An interactive benchmarking suite for multimodal form-filling agents.
\url{https://doi.org/10.48550/arXiv.2506.01520}.

\bibitem{kulkarni2025aerialgym}
Kulkarni, M., Rehberg, W., and Alexis, K. (2025).
Aerial Gym Simulator: A framework for highly parallelized simulation of aerial robots.
\url{https://doi.org/10.48550/arXiv.2503.01471}.

\bibitem{lam2018xview}
Lam, D., Kuzma, R., McGee, K., Dooley, S., Laielli, M., Klaric, M., Bulatov, Y., and McCord, B. (2018).
xView: Objects in context in overhead imagery.
\url{https://doi.org/10.48550/arXiv.1802.07856}.

\bibitem{xia2018dota}
Xia, G.-S., Bai, X., Ding, J., Zhu, Z., Belongie, S., Luo, J., Datcu, M., Pelillo, M., and Zhang, L. (2018).
DOTA: A large-scale dataset for object detection in aerial images.
In {\it Proceedings of the IEEE Conference on Computer Vision and Pattern Recognition}, pp.\ 3974--3983.
\url{https://doi.org/10.1109/CVPR.2018.00418}.

\bibitem{li2020dior}
Li, K., Wan, G., Cheng, G., Meng, L., and Han, J. (2020).
Object detection in optical remote sensing images: A survey and a new benchmark.
{\it ISPRS Journal of Photogrammetry and Remote Sensing}, {\bf 159}:296--307.
\url{https://doi.org/10.1016/j.isprsjprs.2019.11.023}.

\bibitem{sun2022fair1m}
Sun, X., Wang, P., Yan, Z., Xu, F., Wang, R., Diao, W., Chen, J., Li, J., Feng, Y., Xu, T., Weinmann, M., Hinz, S., Wang, C., and Fu, K. (2022).
FAIR1M: A benchmark dataset for fine-grained object recognition in high-resolution remote sensing imagery.
{\it ISPRS Journal of Photogrammetry and Remote Sensing}, {\bf 184}:116--130.
\url{https://doi.org/10.1016/j.isprsjprs.2021.12.004}.

\bibitem{li2024sardet100k}
Li, Y., Li, X., Li, W., Hou, Q., Liu, L., Cheng, M.-M., and Yang, J. (2024).
SARDet-100K: Towards open-source benchmark and toolkit for large-scale SAR object detection.
In {\it Advances in Neural Information Processing Systems}, {\bf 37}:128430--128461.
\url{https://doi.org/10.52202/079017-4079}.

\bibitem{massey2025earthscape}
Massey, M., Munia, N., and Imran, A.-A.-Z. (2025).
EarthScape: A multimodal dataset for surficial geologic mapping and Earth surface analysis.
\url{https://doi.org/10.48550/arXiv.2503.15625}.

\bibitem{stewart2022torchgeo}
Stewart, A.J., Robinson, C., Corley, I.A., Ortiz, A., Ferres, J.M.L., and Banerjee, A. (2022).
TorchGeo: Deep learning with geospatial data.
In {\it Proceedings of the 30th International Conference on Advances in Geographic Information Systems}, pp.\ 1--12.
\url{https://doi.org/10.1145/3557915.3560953}.

\bibitem{mai2023csp}
Mai, G., Lao, N., He, Y., Song, J., and Ermon, S. (2023).
CSP: Self-supervised contrastive spatial pre-training for geospatial-visual representations.
In {\it Proceedings of the 40th International Conference on Machine Learning}.

\bibitem{russwurm2020selfattention}
Ru{\ss}wurm, M., and K{\"o}rner, M. (2020).
Self-attention for raw optical satellite time series classification.
{\it ISPRS Journal of Photogrammetry and Remote Sensing}, {\bf 169}:421--435.
\url{https://doi.org/10.1016/j.isprsjprs.2020.06.006}.

\bibitem{garnot2021panoptic}
Sainte Fare Garnot, V., and Landrieu, L. (2021).
Panoptic segmentation of satellite image time series with convolutional temporal attention networks.
In {\it Proceedings of the IEEE/CVF International Conference on Computer Vision}, pp.\ 4872--4881.
\url{https://doi.org/10.1109/ICCV48922.2021.00483}.

\bibitem{abbas2023xai4eo}
Abbas, A., Linardi, M., Vareille, E., Christophides, V., and Paris, C. (2023).
Towards Explainable AI4EO: An explainable deep learning approach for crop type mapping using satellite images time series.
In {\it IGARSS 2023 -- 2023 IEEE International Geoscience and Remote Sensing Symposium}, pp.\ 1088--1091.
\url{https://doi.org/10.1109/IGARSS52108.2023.10283125}.

\bibitem{christie2018fmow}
Christie, G., Fendley, N., Wilson, J., and Mukherjee, R. (2018).
Functional Map of the World.
In {\it Proceedings of the IEEE/CVF Conference on Computer Vision and Pattern Recognition}.
\url{https://doi.org/10.1109/CVPR.2018.00646}.

\bibitem{gupta2019xbd}
Gupta, R., Goodman, B., Patel, N., Hosfelt, R., Sajeev, S., Heim, E., Doshi, J., Lucas, K., Choset, H., and Gaston, M. (2019).
Creating xBD: A dataset for assessing building damage from satellite imagery.
In {\it Proceedings of the IEEE/CVF Conference on Computer Vision and Pattern Recognition Workshops}, pp.\ 10--17.

\bibitem{wang2025m4sar}
Wang, C., Lu, W., Li, X., Yang, J., and Luo, L. (2025).
M4-SAR: A multi-resolution, multi-polarization, multi-scene, multi-source dataset and benchmark for optical-SAR object detection.
\url{https://doi.org/10.48550/arXiv.2505.10931}.

\bibitem{rouse1974monitoring}
Rouse, J.W., Haas, R.H., Schell, J.A., and Deering, D.W. (1974).
Monitoring vegetation systems in the Great Plains with ERTS.
{\it NASA Special Publication}, {\bf 351}:309--317.

\bibitem{gao1996ndwi}
Gao, B.C. (1996).
NDWI---A normalized difference water index for remote sensing of vegetation liquid water from space.
{\it Remote Sensing of Environment}, {\bf 58}(3):257--266.
\url{https://doi.org/10.1016/S0034-4257(96)00067-3}.

\bibitem{gorelick2017gee}
Gorelick, N., Hancher, M., Dixon, M., Ilyushchenko, S., Thau, D., and Moore, R. (2017).
Google Earth Engine: Planetary-scale geospatial analysis for everyone.
{\it Remote Sensing of Environment}, {\bf 202}:18--27.
\url{https://doi.org/10.1016/j.rse.2017.06.031}.

\bibitem{meta2025sam3}
Carion, N., Gustafson, L., Hu, Y.-T., Debnath, S., Hu, R., Suris, D., Ryali, C., Alwala, K.V., Khedr, H., Huang, A., Lei, J., Ma, T., Guo, B., Kalla, A., Marks, M., Greer, J., Wang, M., Sun, P., R\"adle, R., Afouras, T., et al. (2026).
SAM 3: Segment anything with concepts.
{\it arXiv preprint}, arXiv:2511.16719.
\url{https://doi.org/10.48550/arXiv.2511.16719}.

\bibitem{liu2023groundingdino}
Liu, S., Zeng, Z., Ren, T., Li, F., Zhang, H., Yang, J., Jiang, Q., Li, C., Yang, J., Su, H., Zhu, J., and Zhang, L. (2024).
Grounding DINO: Marrying DINO with grounded pre-training for open-set object detection.
In {\it Computer Vision -- ECCV 2024}, pp.\ 38--55.
\url{https://doi.org/10.1007/978-3-031-72970-6_3}.

\bibitem{openai2026gpt41model}
OpenAI. (2025).
GPT-4.1 model.
OpenAI API documentation.
Available at \url{https://developers.openai.com/api/docs/models/gpt-4.1}.
Accessed: May 1, 2026.

\bibitem{openai2025gpt51}
Singh, A., Fry, A., Perelman, A., Tart, A., Ganesh, A., et al. (2025).
OpenAI GPT-5 System Card.
{\it arXiv preprint}, arXiv:2601.03267.
\url{https://arxiv.org/abs/2601.03267}.

\bibitem{openai2025gptoss120b}
OpenAI, Agarwal, S., Ahmad, L., Ai, J., Altman, S., Applebaum, A., Arbus, E., Arora, R.K., Bai, Y., Baker, B., et al. (2025).
gpt-oss-120b and gpt-oss-20b Model Card.
{\it arXiv preprint}, arXiv:2508.10925.
\url{https://arxiv.org/abs/2508.10925}.

\bibitem{song2024agentbank}
Song, Y., Xiong, W., Zhao, X., Zhu, D., Wu, W., Wang, K., Li, C., Peng, W., and Li, S. (2024).
AgentBank: Towards generalized LLM agents via fine-tuning on 50000+ interaction trajectories.
In {\it Findings of the Association for Computational Linguistics: EMNLP 2024}, pp.\ 2124--2141.
\url{https://doi.org/10.18653/v1/2024.findings-emnlp.116}.

\bibitem{kang2025agentdistill}
Kang, M., Jeong, J., Lee, S., Cho, J., and Hwang, S.J. (2025).
Distilling LLM agent into small models with retrieval and code tools.
arXiv preprint arXiv:2505.17612.
\url{https://doi.org/10.48550/arXiv.2505.17612}.

\bibitem{hu2022lora}
Hu, E.J., Shen, Y., Wallis, P., Allen-Zhu, Z., Li, Y., Wang, S., Wang, L., and Chen, W. (2022).
LoRA: Low-rank adaptation of large language models.
In {\it The Tenth International Conference on Learning Representations}.
\url{https://openreview.net/forum?id=nZeVKeeFYf9}.

\bibitem{bai2025qwen3vl}
Bai, S., Cai, Y., Chen, R., Chen, K., Chen, X., et al. (2025).
Qwen3-VL Technical Report.
{\it arXiv preprint}, arXiv:2511.21631.
\url{https://arxiv.org/abs/2511.21631}.

\bibitem{zhao2025swift}
Zhao, Y., Huang, J., Hu, J., Wang, X., Mao, Y., Zhang, D., Jiang, Z., Wu, Z., Ai, B., Wang, A., Zhou, W., and Chen, Y. (2025).
SWIFT: A scalable lightweight infrastructure for fine-tuning.
{\it Proceedings of the AAAI Conference on Artificial Intelligence}, {\bf 39}(28):29733--29735.
\url{https://doi.org/10.1609/aaai.v39i28.35383}.

\bibitem{comanici2025gemini25}
Comanici, G., Bieber, E., Schaekermann, M., Pasupat, I., Sachdeva, N., et al. (2025).
Gemini 2.5: Pushing the frontier with advanced reasoning, multimodality, long context, and next generation agentic capabilities.
{\it arXiv preprint}, arXiv:2507.06261.
\url{https://arxiv.org/abs/2507.06261}.

\bibitem{cohen1960coefficient}
Cohen, J. (1960).
A coefficient of agreement for nominal scales.
{\it Educational and Psychological Measurement}, {\bf 20}(1):37--46.
\url{https://doi.org/10.1177/001316446002000104}.

\bibitem{chen2021codex}
Chen, M., Tworek, J., Jun, H., Yuan, Q., Pinto, H.P.O., Kaplan, J., Edwards, H., Burda, Y., Joseph, N., Brockman, G., et al. (2021).
Evaluating Large Language Models Trained on Code.
\url{https://doi.org/10.48550/arXiv.2107.03374}.

\bibitem{wang2022selfconsistency}
Wang, X., Wei, J., Schuurmans, D., Le, Q., Chi, E., Narang, S., Chowdhery, A., and Zhou, D. (2023).
Self-consistency improves chain of thought reasoning in language models.
In {\it The Eleventh International Conference on Learning Representations}.
\url{https://openreview.net/forum?id=1PL1NIMMrw}.

\bibitem{wulder2019landsatstatus}
Wulder, M.A., Coops, N.C., Roy, D.P., White, J.C., and Hermosilla, T. (2019).
Current status of Landsat program, science, and applications.
{\it Remote Sensing of Environment}, {\bf 225}:127--147.
\url{https://doi.org/10.1016/j.rse.2019.02.015}.

\bibitem{teknium2024hermes3}
Teknium, R., Quesnelle, J., and Guang, C. (2024).
Hermes 3 Technical Report.
{\it arXiv preprint}, arXiv:2408.11857.
\url{https://doi.org/10.48550/arXiv.2408.11857}.






\end{thebibliography}
\end{document}